\documentclass[11pt,twoside]{article}
\usepackage{informat}
\usepackage{epsfig}
\usepackage[T1]{fontenc}
\usepackage[utf8]{inputenc}
\usepackage{amsmath,amssymb,amsfonts}
\usepackage{graphicx}
\usepackage{booktabs}
\usepackage{multirow}
\usepackage{array}
\usepackage{tabularx}
\usepackage{makecell}
\usepackage{caption}
\usepackage{subcaption}
\usepackage{xcolor}
\usepackage{scalefnt}
\usepackage{adjustbox}
\usepackage{float}
\usepackage{algorithm}
\usepackage{algpseudocode}
\usepackage{placeins}
\usepackage[colorlinks=true,linkcolor=blue,citecolor=blue,urlcolor=blue]{hyperref}

\newcolumntype{C}[1]{>{\centering\arraybackslash}p{#1}}

\graphicspath{{./}}

\begin{document}

\title{Beyond Correlation: Learning Supervised, Sample-Distinct,
  and Eigenimage-Interpretable Representations}

\author{Mojtaba Moattari \\
Independent Researcher\\
ORCID: \href{https://orcid.org/0000-0001-6191-6467}{0000-0001-6191-6467}\\
Moatary.m@gmail.com}
\titleodd{Beyond Correlation}
\authoreven{M. Moattari}
\received{June 9, 2026}

\abstract{Conventional dimensionality reduction methods mainly optimize variance or correlation, leaving statistical dependence, data diversity, contrast, and interpretability under addressed. We propose three new independence criteria for designing supervised and unsupervised dimensionality reduction (DR) methods, aiming to improve feature extraction and representation quality. Our framework combines linear and nonlinear formulations and is evaluated using contrast, classification accuracy, and interpretability measures. The interpretability of eigenfaces helps to effectively summarize dominant class-specific structures and trends within representative images. Evaluated on MNIST and a Gender face dataset for classification and reconstruction, our methods achieve significant improvements in contrast (up to $+$20.1\%), accuracy (up to $+$17.4\%), and interpretability (up to $+$120.0\%) over Principal Component Analysis (PCA), t-distributed Stochastic Neighbor Embedding (t-SNE), Linear Discriminant Analysis (LDA), and Variational Autoencoder (VAE) baselines, while also improving VAE reconstruction performance by 9.5\%. These results suggest a promising direction for interpretable representation learning based on statistical dependence and independence criteria.}
\abstractSi{}

\maketitle

\section{Introduction}
\label{sec:intro}

Dimensionality reduction (DR) is a fundamental problem in machine learning with broad applications in visualization, classification, representation learning, data compression, clustering, and interpretability~\cite{ref3,ref14,ref16,ref19,ref20}. In many practical applications, particularly in image analysis, face recognition, biomedical imaging, and representation learning, the objective is not only to reduce dimensionality but also to preserve meaningful statistical structure, reveal class-specific patterns, and improve the interpretability of learned representations~\cite{ref8,ref9}. Effective DR techniques can therefore improve downstream learning performance while simultaneously exposing latent structures that are difficult to identify in the original feature space~\cite{ref3,ref17}.

Throughout this paper, the terms \emph{dependence} and \emph{independence} are used as shorthand for \emph{statistical dependence} and \emph{statistical independence}, respectively, unless stated otherwise.

Classical DR methods such as PCA~\cite{ref14}, LDA~\cite{ref15}, Independent Component Analysis (ICA)~\cite{ref4,ref5,ref6}, kernel-based approaches~\cite{ref1,ref19}, t-SNE~\cite{ref16}, and VAEs~\cite{ref10,ref13} have demonstrated strong performance in different learning scenarios. PCA emphasizes variance preservation, LDA maximizes class separability, ICA seeks statistically independent latent components, and VAE-based methods learn probabilistic latent representations through reconstruction objectives. Kernel methods extend these approaches to nonlinear settings and can capture more complex structures in data~\cite{ref1,ref19}. More recently, disentangled representation learning and supervised subspace learning methods have shown that interpretable latent factors can improve both discriminative learning and representation quality~\cite{ref8,ref9,ref28,ref30,ref31}. Existing studies have investigated ICA-based independence learning~\cite{ref4,ref5,ref6}, kernel dependence estimation~\cite{ref1,ref2}, supervised PCA methods~\cite{ref3}, and disentangled latent representations~\cite{ref8,ref9,ref10,ref27,ref29}. Despite these advances, existing approaches often struggle to balance between discriminative capability, local structure preservation, representation diversity, contrast enhancement, and interpretability.

A central limitation of many existing methods is that variance-based or correlation-based objectives may fail to capture nonlinear dependencies, fine-grained discriminative structures, and meaningful class-specific subpatterns~\cite{ref1,ref2,ref5}. Linear orthogonality constraints may suppress useful local details and diversity in learned representations~\cite{ref7,ref11,ref12}, while kernel-based approaches often depend heavily on expert-defined kernel functions and may not reliably preserve intrinsic data geometry across different datasets~\cite{ref1,ref19}. Furthermore, many existing DR approaches prioritize global variance preservation or reconstruction quality without explicitly modeling dependence relationships that may be important for variation preservation, downstream classification, visualization, and interpretability~\cite{ref2,ref8,ref9}.

Our main intuition is that effective representations should preserve not only global statistical structure but also class-related contrasts and sample diversity. Based on this intuition, we introduce new dependence and independence criteria for supervised and unsupervised DR. The proposed framework includes both linear and nonlinear formulations and further extends to neural settings through integration with VAEs. By sharing layers between the proposed objectives and VAE architectures, we investigate whether dependence-aware learning can simultaneously improve latent representation quality and reconstruction performance~\cite{ref10,ref13}.

The learned weight matrices of the proposed dependence-based extractors highlight the most discriminative input features, thereby strengthening label supervision. The resulting eigenfaces and eigenimages distill class-specific data into a small set of images that describe characteristic class differences more clearly and naturally than those obtained by competing supervised DR methods. Therefore, they lead to more fault-diagnosed classification pipeline, and more easily interpretable markers or behaviors. In the same way, the proposed dependence-based unsupervised DR methods summarize input images into small number of images that are more locally and globally descriptive of their variations compared to other methods. Projections onto these unsupervised bases therefore provide clearer comparisons and visualizations of variation and differences across high-dimensional samples.

The main contributions of this paper are as follows:

\begin{itemize}
  \item We propose a unified dependence- and independence-based framework for supervised and unsupervised DR that captures discriminative and contrast-enhanced statistical structures beyond conventional correlation-based objectives (RQ1).
  \item We develop an interpretability-oriented analysis framework based on eigenimages and eigenfaces to study class-specific representation behavior, model diagnosis, and feature diversity in reduced-dimensional spaces.
  \item We conduct extensive experiments on MNIST and gender-face datasets and demonstrate consistent improvements over PCA, LDA, ICA variants, t-SNE, and VAE baselines in terms of classification accuracy, contrast quality, reconstruction error, and representation interpretability. The generalization from correlation to dependence (RQ2) is systematically validated through these comparisons.
  \item We introduce linear and neural formulations of the proposed criteria, including layer-sharing mechanisms with VAEs, enabling simultaneous improvement of latent feature quality, interpretability, and reconstruction performance (RQ3).
\end{itemize}

To guide the investigation, we formulate the following three research questions (RQs):

\textbf{RQ1:} How can independence criteria be designed to directly extract contrast-enhanced and label-discriminative features that are simultaneously differentiable and more direct than the standard ICA approach, bypassing the need for a fixed nonlinearity assumption?

\textbf{RQ2:} Does generalizing LDA from correlation to dependence yield higher classification performance and class-specific interpretability scores on supervised datasets?

\textbf{RQ3:} Does performing layer-sharing in the proposed pipeline with a VAE lead to higher K-Nearest Neighbor (KNN) accuracy and class-specific interpretability scores compared with using a VAE or the pipeline alone?

Experimental results demonstrate that the proposed methods consistently outperform existing baselines across multiple evaluation criteria. In particular, the proposed framework improves contrast quality, achieving entropy-score gains of 20.1\% over PCA and 16.2\% over t-SNE. The learned representations also improve classification accuracy by 14.2\% compared with VAE and by 17.4\% compared with LDA. Furthermore, the proposed interpretability framework increases class-related contrasts and visual interpretability scores by up to 120.0\% over Regularized LDA and by 26\% over VAE-based representations. When integrated through layer sharing, the proposed criteria additionally improve VAE reconstruction performance by 9.5\%. These results suggest that dependence-based objectives provide an effective direction for interpretable representation learning, simultaneously advancing quantitative performance and qualitative interpretability through eigenimage and eigenface analysis.

The remainder of this paper is organized as follows. Section~\ref{sec:related} reviews related work and introduces the required background concepts. Section~\ref{sec:methods} presents the proposed dependence criteria and DR algorithms. Section~\ref{sec:eval} reports the experimental setup, quantitative results, and interpretability analysis. Finally, Section~\ref{sec:conclusion} concludes the paper and discusses future research directions.

To ensure full reproducibility, the source code, datasets, and pretrained models used in this study are publicly available at \url{https://github.com/moatary/reproducible_beyond_correlation}. Furthermore, all hyperparameter search spaces and final selected values are detailed in Table~\ref{tab:hyperparams}.

\section{Related Work, Essential Concepts, and Definitions}
\label{sec:related}

\subsection{Definitions}

\textbf{Sample Diversity.}
Sample diversity refers to the degree to which dataset samples are mutually discernible in pairwise distance and local density. For data visualization~\cite{ref16,ref20}, most dataset samples should be mutually discernible in terms of pairwise distances and local density. A supervised DR can prepare data for classification; for unsupervised DR, the contrast between samples should be sufficiently high and diverse so that classifiers or discriminators are not confused in high-density regions.

Diversity is therefore associated with high uncorrelatedness of samples with respect to one another, which corresponds to linear independence in statistics. A feature with high sample diversity can be extracted by a linear transformation that seeks the most uniform statistical density (low data concentration). That viewpoint disentangles the feature from most others, yielding a relatively independent feature with high sample-space diversity. Analyzing each such independent component makes the sample space exponentially less complex to characterize.

\textbf{Feature Disentanglement.}
Disentanglement of features means learning a representation in which each component captures an independent data factor with sufficient variation. A powerful independent component extractor can separate interleaved and complex data patterns into meaningful features, improving data explainability and reducing validation error~\cite{ref8}.

\textbf{Informative Features.}
Informative features convey information about data structure or a prediction target. Because high information capacity or a noisy channel can make diverse features appear as noise or outliers, supervised learning methods add regularization to ensure that more features carry nonzero weight during training. Mutual information between extracted features and labels can further guide feature selection~\cite{ref22}.

\textbf{Span.}
The span of a set of vectors is the set of all possible linear combinations of those vectors.

\textbf{Correlation and Covariance.}
For zero-mean, unit-variance data, the covariance matrix equals the Pearson correlation matrix; throughout this paper ``correlation'' refers to the unnormalized covariance of zero-mean data.

\subsection{Statistical Dependence and Independence}

Maximizing independence is equivalent to minimizing dependence, and vice versa; both formulations yield identical optima. In the following section, we design classifiers based on two objective functions corresponding to dependence and independence.

Although numerous DR methods have been proposed to support downstream supervised or unsupervised models, a unified treatment of dependence-driven criteria across linear and nonlinear settings remains absent from the literature.

Barshan et al.~\cite{ref3} proposed Supervised Principal Component Analysis (SPCA) based on the Hilbert-Schmidt Independence Criterion (HSIC). Given the conceptual overlap with the present work, we examine this method in detail.

First, their proposed matrix to factorize is a variant of the Linear Discriminant:

\begin{align}
\|XL - Y\|^2 &= \mathrm{Tr}\left[(XL - Y)(XL - Y)^\top\right] \notag\\
&= XLL^\top X^\top + YY^\top - 2\,\mathrm{Tr}(XLY^\top) \label{eq:1}
\end{align}

For the relationship between $X$ and $Y$, the first and second right-hand-side terms vanish. The reader is referred to~\cite{ref18} for the derivation.

Therefore $\max\; 2XLY^\top = \min\; {-2XLY^\top}$:

\begin{equation}
= \min\;\mathrm{Tr}\!\left[(XL - Y)(XL - Y)^\top\right] \label{eq:2}
\end{equation}

This shows that SPCA's objective is correlation-based, consistent with PCA and LDA, since all three reduce to a squared-error or trace-norm minimization.

Second, the Power Method and Gram-Schmidt process prevent extracting the next eigenvector without first removing all effects of the previous one (corresponding to the maximum eigenvalue), so the parallel claim does not hold in practice.

Third, SPCA focuses on maximizing dependence on the response variable while ignoring the local structure of the data, and therefore cannot faithfully recover the intrinsic dimensionality.

Fourth, because kernel methods require expert knowledge for kernel selection, they are prone to failure when such expertise is unavailable.

Su et al.~\cite{ref9} proposed a model seeking subspaces that are maximally dependent on the response variable while minimally dependent on each other. Although this yields a diverse set of subspaces each capturing global data structure, the method enforces predefined nonlinearity, and relies on gradient-based optimization in which many bases tend to remain near initialization or vanish entirely. The present work avoids this issue by extracting multiple components sequentially and stopping when the number of iterations exceeds a prescribed limit.

\subsection{Fuzzy Histogram}

A Fuzzy Histogram (FH) is a sequence of real numbers $h(i)$, $i\in\{0,1,\ldots,M{-}1\}$ where $h(i)$ is the frequency of the $i$-th histogram bin. Each membership function is radial and centered at $m(i)$. The FH is computed as:

\begin{equation}
h_i = h_i + \sum_x \mu_I(x)\big|_{k},\quad k\in[x_{\min},x_{\max}] \label{eq:3}
\end{equation}

where $\mu_I(x)$ is the Gaussian fuzzy membership function:

\begin{equation}
\mu_I(x_i) = \frac{1}{\sqrt{2\pi}\,\sigma}\,e^{-\tfrac{(x-x_k)^2}{2\sigma^2}} \label{eq:4}
\end{equation}

The FH handles the inexactness of sample features relative to each bin $h(i)$: values closer to $x_k(i)$ receive higher scores. In this work, because the 2D joint histogram is too sparse to affect gradients effectively, only 1D histograms are used in one of the proposed criteria.

\begin{figure}[htbp]
  \centering
  \includegraphics[width=\columnwidth]{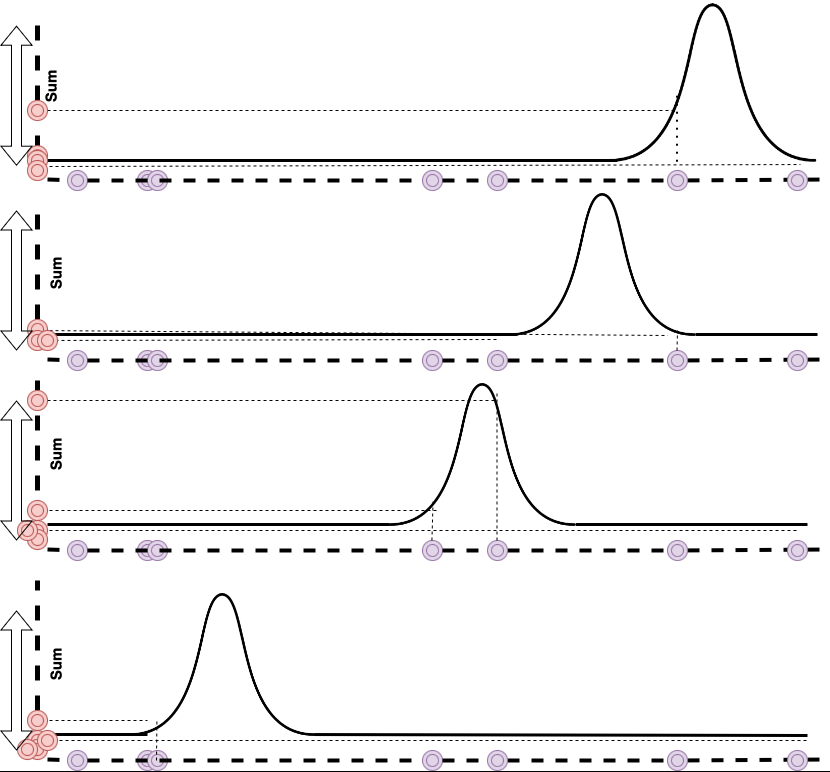}
  \caption{Series of membership functions (per row) for calculating density according to their respective means. Means are drawn randomly.}
  \label{fig:fuzzyhist}
\end{figure}

Figure~\ref{fig:fuzzyhist} illustrates a series of membership functions (one per row) for computing density according to their respective means, which are drawn randomly.

\subsection{Nullspace}

The nullspace of a basis vector $w$ is the span of all orthonormal vectors uncorrelated with $w$. As a result, the inner product of $w$ and $\mathrm{Null}(w) = I - ww^\top$ equals zero, as proved below:

\begin{equation}
\mathrm{Nullspace}(x w^\top) = x(I - ww^\top)
\end{equation}

\begin{align}
\langle x(I-ww^\top),\; xw^\top \rangle &= x(I-ww^\top)wx^\top \notag\\
&= x(w - w(w^\top w))x^\top \notag\\
&= x(w - w)x^\top = 0 \label{eq:5}
\end{align}

Here $\langle\cdot,\cdot\rangle$ denotes the dot product. By minimizing the correlation between a weight-projected feature and all other features (via the nullspace), we learn a high-diversity feature. Specifically, the feature is learned by minimizing the correlation between the feature projection and a random projection of its nullspace:

\begin{align}
w_{\text{optimum}} &: \min_w \langle xw^\top,\;\mathrm{Nullspace}(xw^\top)\cdot a_k\rangle \notag\\
&= \min_w \langle Xw^\top,\;X(I-ww^\top)a_k\rangle \notag\\
&= \min_w \sum_i x_i w^\top \cdot x_i(I-ww^\top)a_k \label{eq:6}
\end{align}

Here $X = [x_1,x_2,\ldots,x_N]^\top$, $a_k$ is a random vector refreshed every $p$ iterations, and $X$, $w$, $a_k$ have dimensions $n\times m$, $m\times 1$, $m\times 1$, respectively.

\subsection{PCA, ICA, and LDA: Background and Relationships}

In this paper, we use the main ideas underlying PCA, ICA, and LDA to design interpretable supervised and unsupervised learning objectives. A deeper understanding of these methods is therefore necessary.

\textbf{Principal Component Analysis (PCA).}
PCA is a linear method that seeks a projection with maximum sample variance, thereby diagonalizing the covariance matrix. In the formulation, a weight vector $w$ is projected onto each zero-mean sample to reduce its dimensionality to one. The resulting scalar is then squared, yielding:

\begin{equation}
w = \arg\max_w\; w^\top(X-\mu_X)^\top(X-\mu_X)w \label{eq:7}
\end{equation}

where $w$ is the projection vector, $X$ is an $m\times n$ matrix of inputs (rows = samples, columns = features), $\mu_X$ is the row-wise sample mean, and $M = Xw$ is the one-dimensional projected data. PCA thus maximizes the variance of the projection.

Because PCA enforces global orthogonality via eigendecomposition rather than nullspace-based decorrelation, it cannot exploit nullspace structure to increase feature contrast. Consequently, it cannot approach decorrelation via the nullspace concept, and in practice it produces lower-contrast representations than the proposed nullspace-based component analysis.

\textbf{Independent Component Analysis (ICA).}
Let $x$ be a random variable taking values in $\mathbb{R}^N$ with probability density $p_{x_i}(u)$. The components of $x$ are mutually independent if and only if

\begin{equation}
p_x(u) = \prod_{i=1}^{N} p_{x_i}(u_i) \label{eq:8}
\end{equation}

Measuring the distance between the two sides of~(\ref{eq:8}) quantifies the degree of independence of a multivariate random variable:

\begin{equation}
d\!\left(p_x,\;\prod_{i=1}^{N} p_{x_i}(u_i)\right) \label{eq:9}
\end{equation}

Key characteristics of ICA are:
\begin{itemize}
  \item It is unsupervised (no label information is used).
  \item Its fixed nonlinearity cannot address all forms of distributional nonlinearity or high-order variable dependencies.
\end{itemize}

\begin{figure}[htbp]
  \centering
  \includegraphics[width=\columnwidth]{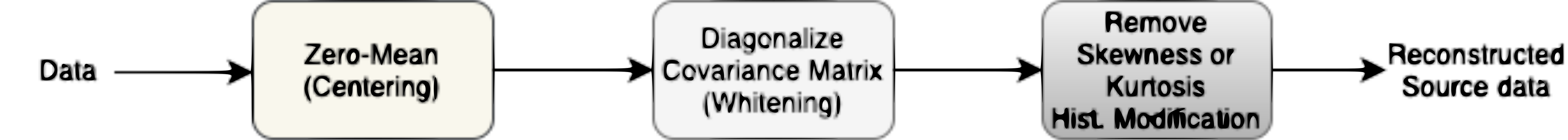}
  \caption{Conventional ICA algorithm procedure.}
  \label{fig:ica}
\end{figure}

\textbf{Regularized Linear Discriminant Analysis (RLDA).}
We seek a regularized variant of LDA. The score function to maximize is:

\begin{align}
\max_w \frac{w^\top C w}{w^\top \Sigma_{C_i} w} &\cong \max_w w^\top C w \notag\\
&\quad \text{s.t.}\; 0 < (w^\top \Sigma_{C_i} w) \ll 1 \label{eq:10}
\end{align}

where $C$ is the whole-data covariance matrix and $\Sigma_{C_i}$ is the sum of within-class covariance matrices. Converting the constraint to a Lagrange multiplier:

\begin{align}
\max_w\; w^\top C w - \lambda\, w^\top \Sigma_{C_i} w &= \max_w w^\top Q w, \notag\\
Q &= C - \lambda\Sigma_{C_i} \label{eq:11}
\end{align}

We extract eigenvectors corresponding to the top $t$ eigenvalues of $Q$. To address ill-posedness, $\lambda$ is fine-tuned and additional matrices are added to $Q$:

\begin{align}
Q_{\text{final}} &= \frac{1{+}\lambda}{(1{-}a)(1{-}b)(1{-}c)}\,C - \lambda\Sigma_{C_i} \notag\\
&\quad + a\,\mathrm{Diag}(\Sigma_{C_i}) + b\,\mathrm{Diag}(C) + c\,I \label{eq:12}
\end{align}

where $I$ is the identity matrix. Inspired by this formulation, we propose a discriminator based on within-class dependence and whole-data independence, where the regularization weights $(\lambda,a,b,c)\in(-1,1)$ balance the contribution of each term.

\subsection{Interpretability}

Model fault diagnosis and behavioral feature feedback are obtained by extracting the most salient features (e.g., those with unusually high or low values, or high information gain). Pixels with unusually high or low values in eigenimages are representative of important features shared across many samples, and certain patterns or textures in eigenimages yield additional insights about data, model, and class behavior.

Su et al.~\cite{ref9} proposed a linear regularization (based on HSIC) to seek maximally dependent subspaces with respect to the response variable while keeping subspaces mutually minimally dependent. Although this yields a wide variety of subspaces each capturing global data structure, the approach avoids deep learning under the assumption that linear models offer superior interpretability. The present work challenges this assumption by demonstrating that the learned weight behavior of a shared layer in a VAE can be meaningfully interpreted for describing a specific target class.

Recent work on concept-based explanations and saliency-based feature attribution~\cite{ref23,ref24} has demonstrated that high-level explanatory features can be extracted from deep networks while maintaining discriminative power. Our eigenimage-based interpretability analysis complements these approaches by providing a class-level, weight-space visualization rather than an input-gradient-based attribution.

\subsection{Independence Criteria in the Literature}

\textbf{Mutual Information Minimization.}
Mutual information (MI) measures how much information one random variable contains about another, quantifying the dependence between two variables. Zero MI indicates statistical independence or near-uniform marginal distributions. Minimizing MI encourages independence among components:

\begin{equation}
I(X;Y) = \sum_{x,y} P_{XY}(x,y)\log\!\frac{P_{XY}(x,y)}{P_X(x)\,P_Y(y)} \label{eq:13}
\end{equation}

\textbf{Kernel PCA~\cite{ref1}.}
Given an $N\times d$ matrix $X$ with $d\geq N$, one can perform eigendecomposition on a kernel matrix rather than on $X$ directly, effectively mapping data to an $N$-dimensional space:

\begin{equation}
K_{ij} = k(x_i, x_j) \label{eq:14}
\end{equation}

The generalized eigenvalue theorem extracts $(W, \Lambda)$ from $K$ via $WK = \Lambda W$, where $W$ ($N\times m$) contains the top $m$ eigenvectors and $\Lambda$ is a diagonal matrix of eigenvalues in descending order.

Kernel PCA can capture nonlinear structure by implicitly mapping data to a higher-dimensional space, effectively performing PCA in that space and uncovering nonlinear relationships that standard PCA would miss. However, because it requires expert knowledge for kernel selection, it is prone to failure in the absence of such expertise, and is therefore not adopted in this work.

\textbf{Hilbert-Schmidt Independence Criterion (HSIC)~\cite{ref2}.}
HSIC is a kernel-based measure of dependence between two random variables. It quantifies independence by embedding probability distributions in a Reproducing Kernel Hilbert Space (RKHS) and computing the Hilbert-Schmidt norm of the cross-covariance operator. The cross-covariance between elements of $F$ and $G$ is:

\begin{equation}
\begin{aligned}
C_{XY} := \mathrm{Cov}(f(x),g(y)) = \mathbb{E}_{x,y}[f(x)g(y)] \\
- \mathbb{E}_x[f(x)]\,\mathbb{E}_y[g(y)] \label{eq:15}
\end{aligned}
\end{equation}

with $F := \mathrm{span}(\{f(x)\})$ and $G := \mathrm{span}(\{g(y)\})$ as RKHSs with feature maps $f$ and $g$, respectively. The HSIC is:

\begin{equation}
\mathrm{HSIC}(X,Y,F,G) := \|C_{xy}\|^2_{HS} \label{eq:16}
\end{equation}

To see why HSIC measures the distance between a joint distribution and the product of marginals, consider substituting $f$ and $g$ with Dirac delta functions:

\begin{equation}
\begin{split}
f(X-x) &= \delta(X-x) \\
&= \lim_{\sigma\to 0}\frac{1}{\sqrt{2\pi}\,\sigma}\,e^{-\tfrac{\|X-x\|^2}{2\sigma^2}},\\
g(Y-y) &= \delta(Y-y) \label{eq:17}
\end{split}
\end{equation}

\begin{multline}
\mathbb{E}_{x,y}[f(x)g(y)] = \iint f(X)g(Y)\,p(X,Y)\,dX\,dY \\
= \iint \delta(X-x)\,\delta(Y-y)\,p(x,y)\,dX\,dY \label{eq:18}
\end{multline}

Since the delta function is zero everywhere except at $x$ and $y$, the constant $p(x,y)$ factors out of the integral:

\begin{multline}
\mathbb{E}_{X,Y}[f(x)g(y)] = p(X{=}x,Y{=}y) \\
\times\iint\delta(X-x)\,\delta(Y-y)\,dX\,dY \label{eq:19}
\end{multline}

Using $\int\delta(u)\,du = 1$:

\begin{equation}
\mathbb{E}_{x,y}[f(x)g(y)] = p(X{=}x,Y{=}y) = p(x,y) \label{eq:20}
\end{equation}

Applying the same reasoning to the marginals yields:
\begin{itemize}
  \item $\mathbb{E}_{x,y}[f(x)g(y)] = p(x,y)$ \quad (joint)
  \item $\mathbb{E}_x[f(x)] = p(x)$ \quad (marginal)
  \item $\mathbb{E}_y[g(y)] = p(y)$ \quad (marginal)
\end{itemize}

Therefore, the probability product rule is recovered: $\mathrm{Cov}(f(x),g(y)) = p(x,y) - p(x)p(y)$, which serves as a measure of dependence.

A practical empirical estimate of HSIC given $E=\{(x_1,y_1),\ldots,(x_n,y_n)\}\subseteq X\times Y$ is:

\begin{equation}
\mathrm{HSIC}(Z,F,G) := (n-1)^{-2}\,\mathrm{tr}(KHLH) \label{eq:21}
\end{equation}

where $H,K,L \in \mathbb{R}^{n\times n}$, $K_{ij} := k(x_i,x_j)$, $L_{ij} := l(y_i,y_j)$, and $H_{ij} := I - n^{-1}\mathbf{e}\mathbf{e}^\top$ is the centering matrix.

Although HSIC can capture nonlinear structure when the chosen kernel increases contrast between samples, its reliance on expert knowledge for the selection of $f$, $g$, or the kernel makes it unsuitable for the proposed methods.

\subsection{Recent Advances in Dependence-Based and Disentangled Representation Learning}

Several recent studies have further explored statistical dependence, self-supervised objectives, and disentangled representations as foundations for DR, motivating the present work.

Chen and Ji~\cite{ref27} proposed a self-supervised DR framework that maximizes feature-label mutual dependence without requiring explicit label supervision during projection, demonstrating that dependence-based objectives generalize well beyond kernel methods to neural feature extractors. Their work underscores the importance of measuring dependence directly rather than relying on proxy objectives such as reconstruction error. 

Federici et al.~\cite{ref28} analyzed the information-theoretic bounds of disentangled representations and showed that minimizing mutual information between latent components while preserving label-relevant information produces more interpretable and compact representations than variance-based methods, establishing a formal connection between disentanglement and independence criteria. 

Zbontar et al.~\cite{ref29} introduced Barlow Twins, a self-supervised learning method that enforces redundancy reduction by minimizing off-diagonal elements of the cross-correlation matrix of twin-network embeddings. This approach operationalizes statistical independence between representation dimensions as a training objective, directly relating to the decorrelation criteria proposed in the present work. 

Lee et al.~\cite{ref30} proposed a supervised subspace learning method using conditional independence criteria to extract class-discriminative features for medical image analysis. Their results confirm that conditioning on class labels when estimating independence measures yields stronger discriminative representations than global independence objectives, consistent with the within-class dependence strategy adopted here. 

Wang et al.~\cite{ref31} demonstrated that disentangled latent representations learned via variational objectives with structured independence constraints substantially improve downstream classification and visualization quality in high-dimensional biomedical data. Their findings support the use of combined VAE and independence objectives, as investigated through the layer-sharing mechanism in the present work. 

\section{Proposed Methods}
\label{sec:methods}

The independence criteria reviewed in Section~\ref{sec:related} are repurposed here as dependence criteria by negating their objectives: minimizing independence is equivalent to maximizing dependence. Inspired by the RLDA formulation, we design supervised DR methods that explicitly optimize dependence within classes while enforcing independence across the whole dataset.

\subsection{Proposed Independence Criteria}

We propose three independence criteria and apply them within the supervised and unsupervised models introduced below.

\textbf{1. Nullspace-Based Decorrelation (Linear Independence).}

\begin{figure}[htbp]
  \centering
  \includegraphics[width=\columnwidth]{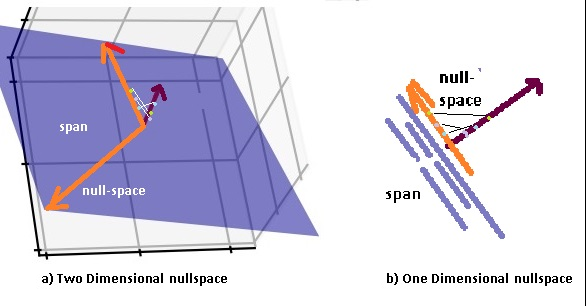}
  \caption{Nullspace of a weight vector (red) and its spans (blue) in 2D and 3D. Each dot is shown with the same color in both the feature space and the random nullspace span. Variables in the dots are interwoven and not yet disentangled, as the ordering in the projection space differs from that in the nullspace span.}
  \label{fig:nullspace}
\end{figure}

The objective is to find the optimal $w$ that minimizes the correlation between two projections across all samples:

\begin{align}
&\min\;\mathbb{E}\!\left[(w^\top x),\;(r\,\mathrm{Null}(w)\,x)\right] \notag\\
&= \frac{1}{N}\sum_i (w^\top X_i)\,(r\,(I-ww^\top)X_i) \label{eq:22}
\end{align}

where $r$ is a random vector, $w$ is a column vector, $X$ is $N\times d$, and $N$ is the number of samples. Although this criterion minimizes only linear dependence, applying it within a neural network allows nonlinear feature maps, so that decorrelated components can be mined across most features in a nonlinear sense.

Note that the decorrelation here applies to features, not to samples directly. The objective creates two views---\emph{(i)} the projection of samples onto a basis $w$, and \emph{(ii)} the projection onto a random span of the nullspace---to increase the feature diversity of the resulting sample representation. Algorithm~\ref{pseudo:1} is related to this subsection.

\begin{algorithm}[t]
\caption{Minimizing Correlation of Weight Parameter and its Nullspace's Random View}
\label{pseudo:1}
\begin{algorithmic}[1]
\Require \texttt{wParameter} (column vector), $\langle\cdot,\cdot\rangle$ (inner product)
\Ensure \texttt{wParameter}
\State $S \leftarrow \texttt{Data\_Sampler}(\text{`Uniform'}, \text{batch\_size}=128)$
\While{$\text{iter} \leq \text{MaxIter}$}
  \State $\texttt{SPS} \leftarrow \langle \texttt{wParameter},\; S^\top\rangle$
  \State $\texttt{rNSS} \leftarrow \langle \texttt{randVec},\; (I - \texttt{wParam}\cdot\texttt{wParam}^\top)\cdot S^\top\rangle$
  \State $\texttt{wParameter} \mathrel{-}= \nabla_w \langle \texttt{SPS},\;\texttt{rNSS}^\top\rangle$
\EndWhile
\State \Return \texttt{wParameter}
\end{algorithmic}
\end{algorithm}

\textbf{2. Probability Product Rule Criterion with Differentiable Fuzzy Histogram.} 

Independent features yield rich, disentangled, and highly diverse representations that contrast samples more effectively than conventional ICA. Existing ICA algorithms, however, do not apply the probability product rule directly as their independence objective. The probability product rule states that two random variables $a$ and $b$ are independent if and only if their joint density equals the product of their marginal densities; the deviation from this equality therefore measures statistical dependence:

\begin{align}
L &= \sum\!\left(P_A(a)\,P_B(b) - P_{AB}(a,b)\right), \notag\\
&\quad a = \langle w,X\rangle,\quad b = r(I-ww^\top)X \label{eq:23}
\end{align}

Minimizing $L$ enforces independence between $a$ and $b$; maximizing $L$ enforces dependence. Following~(\ref{eq:22}), we construct fuzzy histograms of $P_A(a)$, $P_B(b)$, and $P_{AB}(a,b)$ with preserved computational graph gradients (see Section~\ref{sec:related}).

Indirect cost functions such as non-Gaussianity, entropy, or correlation provide only necessary, not sufficient, conditions for independence. By contrast, the probability product rule criterion directly measures the gap between the joint distribution and the product of marginals, providing a sufficient criterion for independence.  Algorithm~\ref{pseudo:2} is related to this subsection.

\begin{algorithm}[t]
\caption{Minimizing Dependence of Weight Parameter and its Nullspace's Random View}
\label{pseudo:2}
\begin{algorithmic}[1]
\Require \texttt{wParameter} (column vector), $\langle\cdot,\cdot\rangle$ (inner product)
\Ensure \texttt{wParameterOptimal}
\State $S \leftarrow \texttt{Data\_Sampler}(\text{`Uniform'}, \text{batch\_size}=128)$
\While{$\text{iter}\leq\text{MaxIter}$}
  \State $\texttt{SPS} \leftarrow \langle \texttt{wParameter},S^\top\rangle$
  \State $\texttt{rNSS} \leftarrow \langle \texttt{randVec},(I - \texttt{wParam}\cdot\texttt{wParam}^\top)\cdot S^\top\rangle$
  \State $H_1 \leftarrow \texttt{Hist1D}(\texttt{SPS})$
  \State $H_2 \leftarrow \texttt{Hist1D}(\texttt{rNSS})$
  \State $H_{12} \leftarrow \texttt{Hist2D}(\texttt{zip}(\texttt{SPS},\texttt{rNSS}))$
  \State $\texttt{wParameter} \mathrel{-}= \nabla_w \left(\texttt{outer}(H_1,H_2) - H_{12}\right)^2$
\EndWhile
\State \Return \texttt{wParameter}
\end{algorithmic}
\end{algorithm}

\textbf{3. Maximum Entropy of Marginal Histograms.}

\begin{figure}[htbp]
  \centering
  \includegraphics[width=\columnwidth]{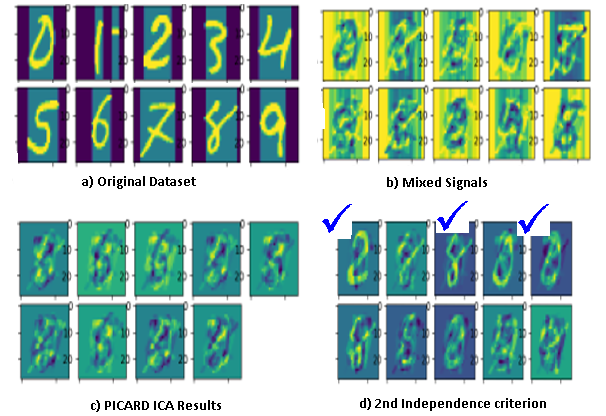}
  \caption{The proposed ICA based on the second independence criterion outperforms the PICARD algorithm on the image unmixing task.}
  \label{fig:hist1d2d}
\end{figure}

Because 2D histograms are very sparse and their gradients negligible, setting $P_{AB}(a,b)=0$ in~(\ref{eq:23}) reduces the objective to maximizing the marginal entropies of $P_A(a)$ and $P_B(b)$, driving each distribution toward uniformity. Figure~\ref{fig:hist1d2d} illustrates the density sparsity difference between 1D and 2D histograms. The resulting cost function is:

\begin{align}
w &= \arg\max_w\;-\mathbb{E}[\log P_A(a)] - \mathbb{E}[\log P_B(b)] \notag\\
&= -\frac{1}{N}\sum_i \log P_X(w^\top X_i) - \frac{1}{N}\sum_i \log P_X\!\left(r(I-ww^\top)X_i\right) \label{eq:24}
\end{align}

Algorithm~\ref{pseudo:3} is related to this subsection.

\begin{algorithm}[t]
\caption{Maximizing Entropy of Marginal Histograms}
\label{pseudo:3}
\begin{algorithmic}[1]
\Require \texttt{wParameter} (column vector), $\langle\cdot,\cdot\rangle$ (inner product)
\Ensure \texttt{wParameterOptimal}
\State $S \leftarrow \texttt{Data\_Sampler}(\text{`Uniform'}, \text{batch\_size}=128)$
\While{$\text{iter}\leq\text{MaxIter}$}
  \State $\texttt{SPS} \leftarrow \langle \texttt{wParameter},S^\top\rangle$
  \State $\texttt{rNSS} \leftarrow \langle \texttt{randVec},(I - \texttt{wParam}\cdot\texttt{wParam}^\top)\cdot S^\top\rangle$
  \State $\texttt{wParameter} \mathrel{-}= \nabla_w\!\Big[-\tfrac{1}{N}\sum_i \log P_X(w^\top X_i)$
  \Statex \hspace{4em} $- \tfrac{1}{N}\sum_i \log P_X\!\left(r(I-ww^\top)X_i\right)\Big]$
\EndWhile
\State \Return \texttt{wParameter}
\end{algorithmic}
\end{algorithm}

\subsection{Relationship Between Proposed and Conventional DRs}

To understand the role of conventional DR methods in improving classifier performance, we compare accuracy and interpretability power using a process analogous to the second proposed independence criterion.

The second independence criterion is also used to design a direct ICA variant that applies the probability product rule criterion (Eq.~(\ref{eq:23})) to unmix one component at a time. This addresses scenarios in which conventional ICA methods fail because they either relax the independence constraint to uncorrelatedness (PCA, kurtosis-based ICA) or impose a predefined fixed distribution. The ``cocktail party problem''---where mixed audio sources must be separated---provides an intuitive analogy: here, dataset images are linearly combined and must be unmixed. Picard~\cite{ref7} is the state-of-the-art maximum-likelihood ICA algorithm against which we compare.

Source localization via Electroencephalography (EEG) provides a further evaluation of the first independence criterion. The EEG data originate from a speech-imagery BCI experiment developed by Rostami et al.~\cite{ref21}, recorded from 16 channels over 6 subjects (ages 23--30), each performing imagination of five vowel sounds (180 trials per subject, 512 Hz, 4 s per trial). A 5-fold cross-validation protocol is used (Algorithm~\ref{pseudo:4}).

\begin{algorithm}[t]
\caption{Accuracy of Speech Imagery for Brain-Computer Interface (BCI)}
\label{pseudo:4}
\begin{algorithmic}[1]
\Require Data ($N\times d$), $N$ samples, $d$ variables
\Ensure Accuracy of classified independent components
\State Randomly partition trials into 5 equal folds.
\State Zero-mean and unit-variance normalize each channel.
\For{each fold}
  \State Use remaining folds as training data and selected fold as test set.
  \State Decimate by factor 8; bandpass filter (5th-order Chebyshev, 3--30 Hz).
  \State Learn ICA projector on training data.
  \State Project train and test data through the ICA demixing matrix.
  \State Extract the top 16 significant components from both splits.
  \State Classify with linear-kernel Support Vector Machine (SVM) (LibSVM); record test accuracy.
\EndFor
\State Average accuracy over 5 folds.
\end{algorithmic}
\end{algorithm}

\subsection{Proposed Learning Algorithms}

We propose five model variants to improve unsupervised and supervised learning: a linear unsupervised learner, a linear supervised learner, a nonlinear neural unsupervised learner, a nonlinear neural supervised learner, and a VAE layer-sharing scheme.

\textbf{Linear Space-Nullspace Decorrelative Components (LSNDC).}
This method minimizes the correlation between a feature projection and a random span of its nullspace (Algorithm~\ref{pseudo:1}), ensuring high sample diversity and high entropy in the extracted feature. This method is a linear version of independence between projection and nullspace of projection. The reason is that linear correlation changes Equation~\ref{eq:1} to $\mathbb{E}[xy]-\mathbb{E}[x]\mathbb{E}[y]$.

To cover multiple basins of attraction, the weight vector (\texttt{wParameter}) is randomly initialized many times and each initialization is updated until convergence; the final positions are retained as candidate components. To ensure all data dependencies and interactions are compressed in the reduced dimensionality, we initialize the weight (\texttt{wParameter} in Algorithm~\ref{pseudo:1}) by random vectors many times, and save its final position (basin of attraction) after adequate gradient updates. We can use a metaheuristic optimization to find the optimum number of weight parameters, and set its cost to either decorrelation with nullspace or to the probability product rule criterion~(\ref{eq:23}). For this linear variant, independent components (e.g., face subparts or nonlinear facial textures) are not expected unless a kernel or nonlinear feature map is applied.

\textbf{Linear Independence-Based Component Analysis (LIBCA).}
This method uses Algorithm~\ref{pseudo:2} or~\ref{pseudo:3}. The weight parameter is randomly initialized multiple times and all copies are updated via backpropagation, applying the probability product rule criterion rather than correlation to achieve independence. The rationale is that this direct independence criterion may reduce the reliance on a neural network when the data exhibit nonlinear dependencies.

\textbf{Whole-Data Decorrelation Within-Class Correlation (WDDWCC).}
A projection is made correlated with its nullspace (gradient ascent on Algorithm~\ref{pseudo:1}) for within-class samples, while the whole-data term decorrelates (gradient descent on Algorithm~\ref{pseudo:1}):

\begin{align}
L = &\min\;\frac{1}{N}\sum_i (w^\top X_i)(r(I-ww^\top)X_i) \notag\\
&-\sum_j \frac{1}{N}\sum_i (w^\top X[\mathrm{target}{=}j][i])(r(I-ww^\top)X[\mathrm{target}{=}j][i]) \label{eq:25}
\end{align}

\textbf{Whole-Data Independence Within-Class Dependence (WDIWCD).}
The projection is made dependent on its nullspace for within-class samples (Algorithm~\ref{pseudo:2} or~\ref{pseudo:3}), while the whole-data term minimizes dependence:

\begin{align}
L &= \|P_A(a)\,P_B(b) - P_{AB}(a,b)\|^2 \notag\\
&\quad - \sum_j \frac{1}{|\mathrm{unique}(\mathrm{target})|}\ \notag\\
&\quad \times \|P_A(a_{t=j})\,P_B(b_{t=j}) - P_{AB}(a_{t=j},b_{t=j})\|^2 \notag\\
&a = \langle w,X^\top\rangle,\quad b = r(I-ww^\top)X^\top \label{eq:26}
\end{align}

\begin{figure}[htbp]
  \centering
  \includegraphics[width=\columnwidth]{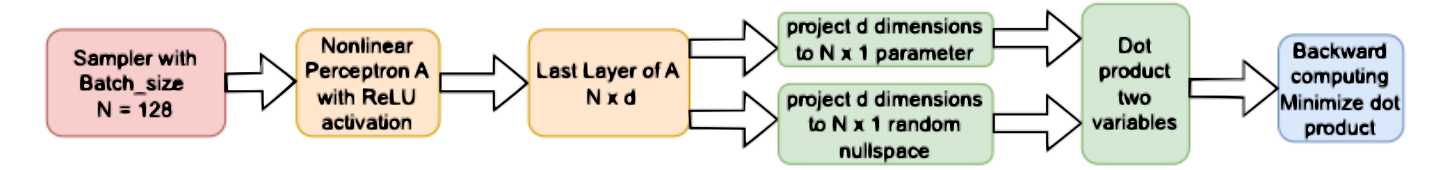}
  \caption{Schematics of Nonlinear Neural Space-Nullspace Decorrelative Components (NNSNDC).}
  \label{fig:nnsndc}
\end{figure}

\begin{figure}[htbp]
  \centering
  \includegraphics[width=\columnwidth]{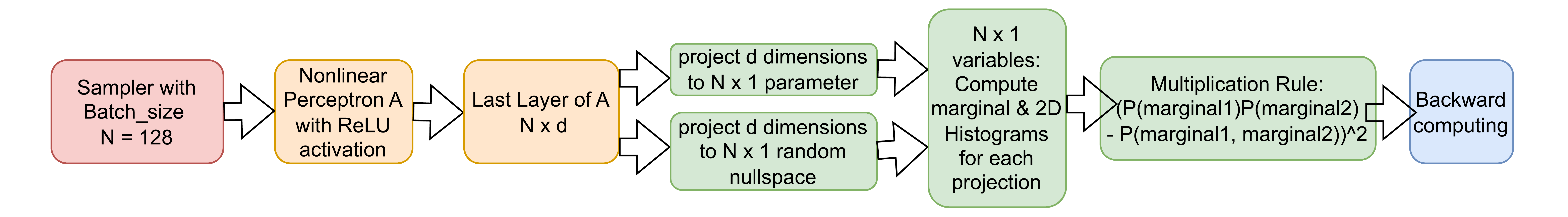}
  \caption{Schematics of Nonlinear Neural Independence-Based Component Analysis (NNIBCA).}
  \label{fig:nnibca}
\end{figure}

\begin{figure}[htbp]
  \centering
  \includegraphics[width=\columnwidth]{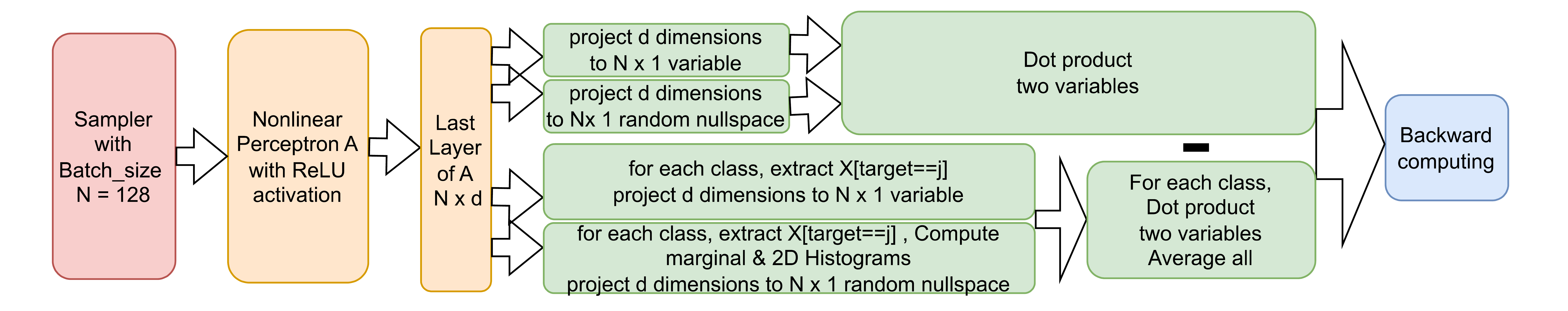}
  \caption{Schematics of Neural Whole-Data Decorrelation Within-Class Correlation (NWDDWCC).}
  \label{fig:NWDDWCC}
\end{figure}

\begin{figure}[htbp]
  \centering
  \includegraphics[width=\columnwidth]{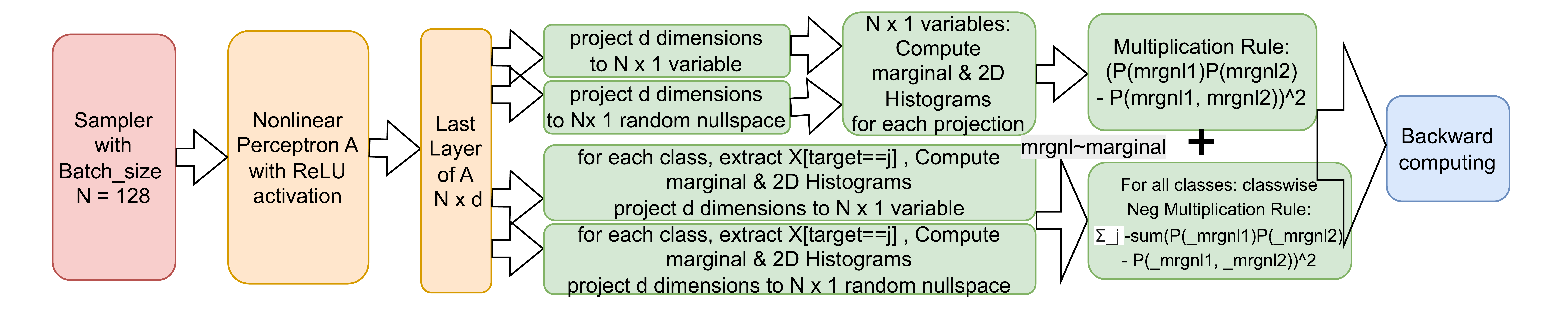}
  \caption{Schematics of Neural Whole-Data Independence Within-Class Dependence (NWDIWCD).}
  \label{fig:nWDIWCD}
\end{figure}

\textbf{Nonlinear Neural Models.}
In this subsection, the proposed models are respectively Nonlinear Neural Space-Nullspace Decorrelative Components (NNSNDC) (Figure~\ref{fig:nnsndc}), Nonlinear Neural Independence-Based Component Analysis (NNIBCA) (Figure~\ref{fig:nnibca}), Neural Whole-Data Decorrelation Within-Class Correlation (NWDDWCC) (Figure~\ref{fig:NWDDWCC}), and Neural Whole-Data Independence Within-Class Dependence (NWDIWCD) (Figure~\ref{fig:nWDIWCD}).

Neural unsupervised learners (e.g., VAE~\cite{ref13}, PCA Network~\cite{ref17}) capture nonlinear dependencies. The neural variants---NNSNDC and NNIBCA---extend LSNDC and LIBCA by passing features through a multilayer perceptron before computing the decorrelation or independence loss. Specifically, the MLP used in NNSNDC and NNIBCA consists of two fully connected layers (input $\to$ 256 units with ReLU activation $\to$ latent dimension), making it a two-layer perceptron. The VAE architecture comprises six layers in total: a three-layer encoder (input $\to$ 512 units $\to$ 256 units $\to$ latent dimension, all with ReLU) and a symmetric three-layer decoder (latent dimension $\to$ 256 units $\to$ 512 units $\to$ output, with sigmoid on the final layer). The supervised neural variants NWDDWCC and NWDIWCD share the same two-layer MLP backbone, to which a second class-conditioned cost term is added. To implement nonlinear decorrelation, a dot-product operation (of the projected feature onto its nullspace) is added to the two-layer perceptron operating over the variable and its randomly spanned nullspace. The resulting sample relationship plots exhibit higher contrast than those of mere PCA. Moreover, we have tested the hypothesis that feature projections with high contrast improve supervised learning performance.

The supervised neural variants---NWDDWCC and NWDIWCD---add a second cost term that operates on each class separately.

In NWDDWCC, the within-class correlation of features with their nullspace pulls same-class samples together. As evident in Figure~\ref{fig:NWDDWCC}, the second block (right side) shows that the global dot product is subtracted from the within-class dot product. Correlation of within-class features with their nullspace makes the extracted features of the class entangled and relevant to each other, allowing samples of the class to cluster near the class center. This makes them far from other classes.

In NWDIWCD, the within-class dependence quantified by the probability product rule criterion makes class features more similar to one another, promoting invariant features and improving KNN classification. Dependence---in terms of the magnitude of the deviation of the joint PDF from the product of marginals ---should be high for within-class samples and low for the whole dataset. This way, global data can yield a feature that captures data variation more readily than correlation-based models. When the within-class product rule deviation is large, the features become more mutually dependent and therefore cluster more easily together, mimicking similar patterns or yielding invariant features. Consequently, after training this model, KNN is expected to achieve a high classification score.

\textbf{Layer Sharing with VAE.}

\begin{figure}[htbp]
  \centering
  \includegraphics[width=\columnwidth]{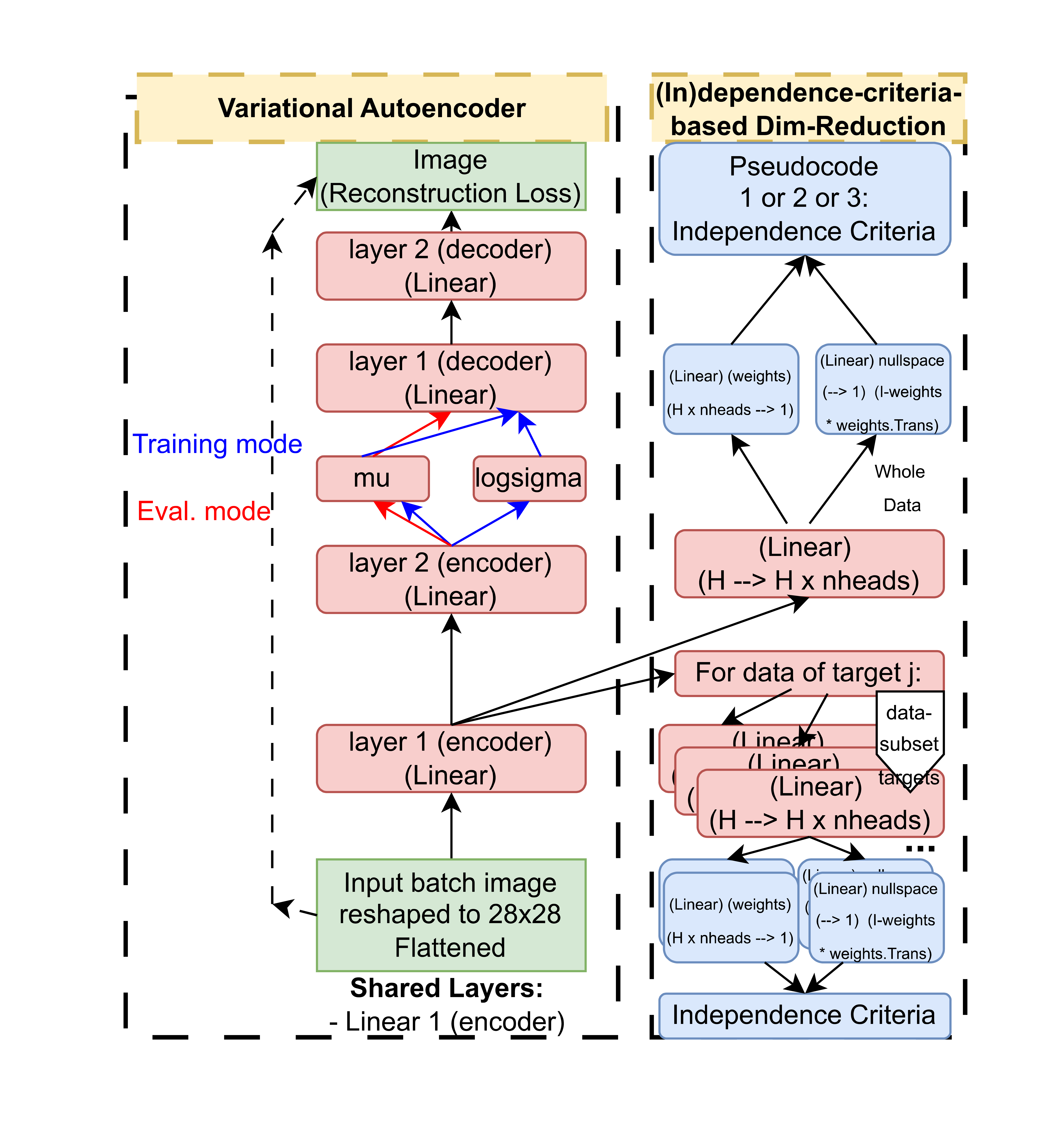}
  \caption{Schematics of layer sharing with VAE: sharing NWDIWCD with the first layer of a Variational Autoencoder.}
  \label{fig:layershare}
\end{figure}

\begin{figure}[htbp]
  \centering
  \includegraphics[width=\columnwidth]{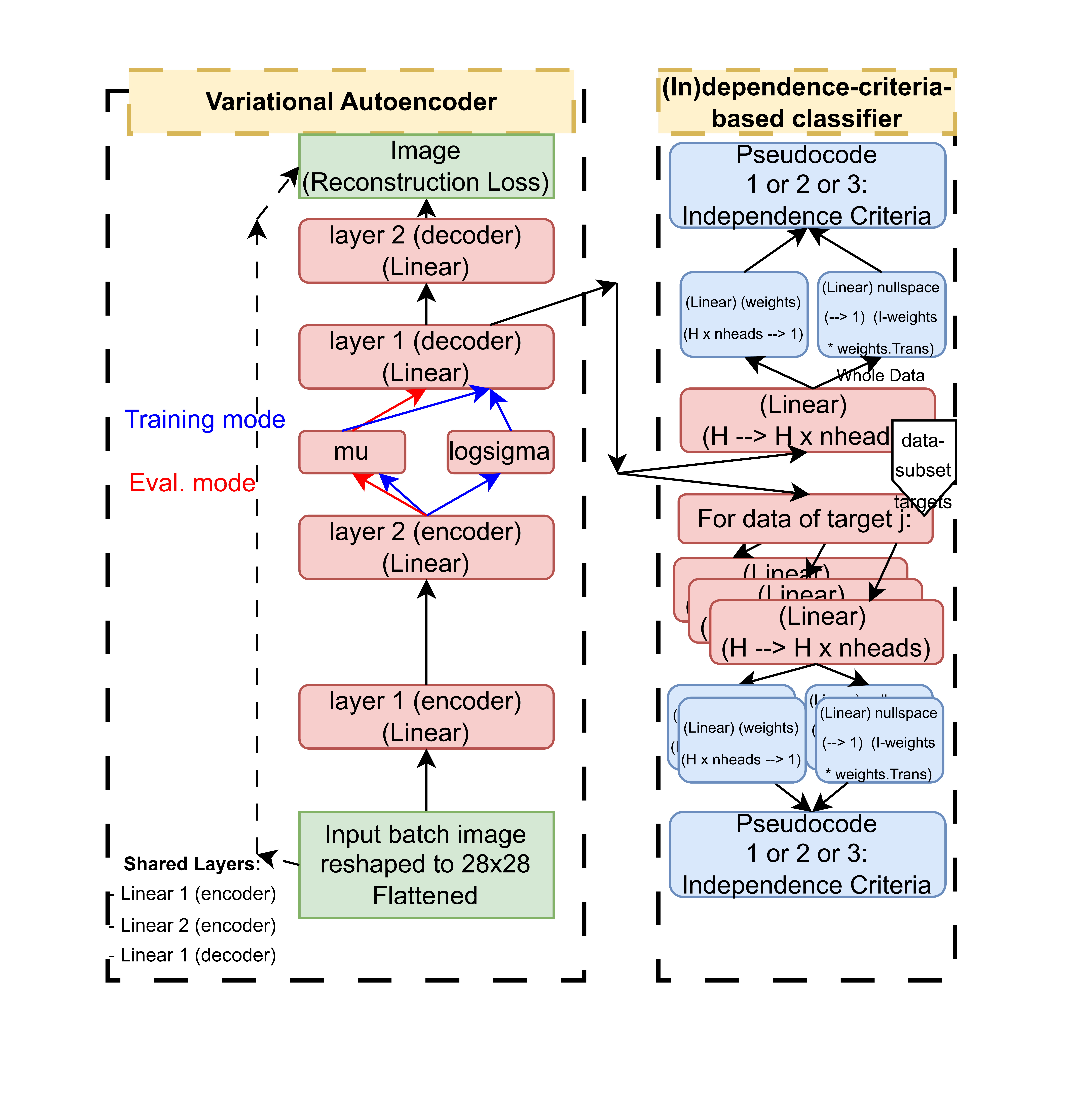}
  \caption{Schematics of most-layers sharing with VAE: all layers except the last are shared with the Variational Autoencoder.}
  \label{fig:mostsharelayers}
\end{figure}

In this subsection, the proposed model is Variational AutoEncoder Neural Whole-Data Independence Within-Class Dependence (VAE-NWDIWCD) (Figure~\ref{fig:layershare} and ~\ref{fig:mostsharelayers}),

By sharing the first layer of a VAE with NWDIWCD, both fine-grained (discriminative) and coarse-grained (global descriptive) patterns can inform the VAE's subsequent layers. The combined loss is:

\begin{align}
L_{\text{VAE-WDI-WCD}} &= (1-a)(1-b)\cdot\mathrm{Recon}_{\text{VAE}} \notag\\
&\quad + b\cdot\mathrm{KLD}_{\text{VAE}} \notag\\
&\quad + a\!\left((1-c)\cdot\text{within\_class} + c\cdot\text{whole\_data}\right) \label{eq:28}
\end{align}

The scalar $c$ controls the balance between supervised (discriminative) and unsupervised (holistic) information flowing into the shared layer, enabling fine-tuning of this trade-off.

\begin{table}[htbp]
\caption{Hyperparameter search space and final tuned values for the proposed models.}
\label{tab:hyperparams}
\resizebox{\columnwidth}{!}{%
\begin{tabular}{|l|l|l|c|}
\hline
\textbf{Model} & \textbf{Parameter} & \textbf{Search Space} & \textbf{Value} \\
\hline
\multirow{5}{*}{\makecell[l]{WDDWCC}}
  & LR & $\{0.001,0.01,0.05,0.1\}$ & 0.01 \\
  & Batch & $\{64,128,256\}$ & 128 \\
  & MaxIter & $\{100,500,1000,2000\}$ & 1000 \\
  & $\lambda$ & $[-1,1]$ & 0.5 \\
  & Components & $\{4,8,16,32\}$ & 16 \\
\hline
\multirow{5}{*}{\makecell[l]{WDIWCD}}
  & LR & $\{0.001,0.01,0.05,0.1\}$ & 0.01 \\
  & Batch & $\{64,128,256\}$ & 128 \\
  & MaxIter & $\{100,500,1000,2000\}$ & 1000 \\
  & Hist.\ bins $M$ & $\{8,16,32,64\}$ & 32 \\
  & Components & $\{4,8,16,32\}$ & 16 \\
\hline
\multirow{6}{*}{\makecell[l]{VAE}}
  & Latent dim & $\{8,16,32,64\}$ & 16 \\
  & LR & $\{0.0001,0.001,0.01\}$ & 0.001 \\
  & Batch & $\{32,64,128\}$ & 64 \\
  & $\beta$ (KLD) & $\{0.1,0.5,1.0,2.0\}$ & 1.0 \\
  & Enc.\ depth & $\{2,3,4\}$ & 3 \\
  & Epochs & $\{50,100,200\}$ & 100 \\
\hline
\multirow{8}{*}{\makecell[l]{VAE+WDIWCD}}
  & $a$ (prop.\ wt.) & $[0,1]$ & 0.80 \\
  & $b$ (KLD wt.) & $[0,1]$ & 0.48 \\
  & $c$ (whole-data) & $[0,1]$ & 0.87 \\
  & Latent dim & $\{8,16,32,64\}$ & 16 \\
  & LR & $\{0.0001,0.001,0.01\}$ & 0.001 \\
  & Batch & $\{32,64,128\}$ & 64 \\
  & Hist.\ bins $M$ & $\{8,16,32,64\}$ & 32 \\
  & Epochs & $\{50,100,200\}$ & 100 \\
\hline
\end{tabular}}
\end{table}

\section{Evaluation, Analysis, and Interpretability Results}
\label{sec:eval}

\subsection{Datasets and Experimental Protocol}
\label{sec:datasets}

Small datasets are used intentionally, because several DR methods require eigenvalue decomposition and the number of feature dimensions may exceed the batch size---a challenge for eigenvalue decomposition that does not affect neural network methods.

\textit{MNIST.} The MNIST database contains 60,000 training and 10,000 test images of handwritten digits, normalized to $28\times28$ pixels with anti-aliasing grayscale levels~\cite{ref25}.

\textit{Gender.} A dataset of male and female faces, preprocessed via facial landmark triangulation, warping to a common landmark set, histogram equalization, rotation correction, and resizing to $28\times28$. It contains 2,150 training images and 350 test images. Representative samples from both datasets are shown in the last row of Table~\ref{tab:table2}.

\begin{table}[htbp]
\caption{Summary of datasets used in experiments. All images are resized to $28\times28$ grayscale. \textit{Classes}: number of output classes used for KNN evaluation. \textit{Preprocessing}: key steps applied before feature extraction.}
\label{tab:datasets}
\resizebox{\columnwidth}{!}{%
\begin{tabular}{|l|c|c|c|c|l|}
\hline
\textbf{Dataset} & \textbf{Train} & \textbf{Test} & \textbf{Classes} & \textbf{Size} & \textbf{Preprocessing} \\
\hline
MNIST & 60{,}000 & 10{,}000 & 10 (2 for Acc2) & $28{\times}28$ & Anti-aliasing grayscale normalisation \\
\hline
Gender (faces) & 2{,}150 & 350 & 2 & $28{\times}28$ & Landmark triangulation, histogram equalisation, rotation correction \\
\hline
\end{tabular}}
\end{table} 

All accuracy results in Tables~\ref{tab:table3} and~\ref{tab:table8} are reported as mean~$\pm$~standard deviation over five independent runs. Statistical significance is assessed with a paired $t$-test (significance level $\alpha=0.05$) between each proposed method and the corresponding baseline; the symbol `$\wedge$' in the tables marks results that are statistically significant under this test. 

\begin{table*}[!ht]
\centering
\caption{Comparing PCA with the proposed linear unsupervised method. Each eigenface results from a different random initialization of the projection weight vector. The proposed method yields more natural, class-summarized eigenfaces.}
\label{tab:table2}
\small
\setlength{\tabcolsep}{3pt}
\resizebox{\textwidth}{!}{%
\begin{tabular}{|l|*{5}{c|}*{5}{c|}}
\hline
\textbf{Gender DS} & male & male & male & male & male & female & female & female & female & female \\
\hline

\makecell[l]{\textbf{Null-space Decorrelation} \\ \textbf{(Pseudocode1)} \\ \textbf{(Proposed)}} &
\includegraphics[width=0.638cm,height=0.734cm]{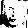} &
\includegraphics[width=0.734cm,height=0.734cm]{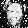} &
\includegraphics[width=0.734cm,height=0.734cm]{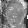} &
\includegraphics[width=0.734cm,height=0.734cm]{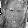} &
\includegraphics[width=0.734cm,height=0.734cm]{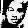} &
\includegraphics[width=0.734cm,height=0.734cm]{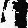} &
\includegraphics[width=0.734cm,height=0.734cm]{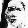} &
\includegraphics[width=0.734cm,height=0.734cm]{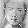} &
\includegraphics[width=0.734cm,height=0.734cm]{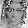} &
\includegraphics[width=0.734cm,height=0.734cm]{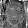} \\
\hline

\textbf{PCA} &
\includegraphics[width=0.704cm,height=0.764cm]{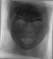} &
\includegraphics[width=0.681cm,height=0.760cm]{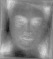} &
\includegraphics[width=0.734cm,height=0.811cm]{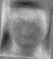} &
\includegraphics[width=0.760cm,height=0.838cm]{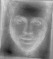} &
\includegraphics[width=0.734cm,height=0.811cm]{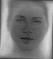} &
\includegraphics[width=0.785cm,height=0.864cm]{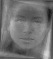} &
\includegraphics[width=0.785cm,height=0.864cm]{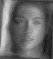} &
\includegraphics[width=0.785cm,height=0.864cm]{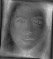} &
\includegraphics[width=0.785cm,height=0.864cm]{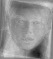} &
\includegraphics[width=0.785cm,height=0.864cm]{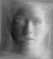} \\
\hline

\textbf{Dataset} &
\includegraphics[width=0.891cm,height=0.995cm]{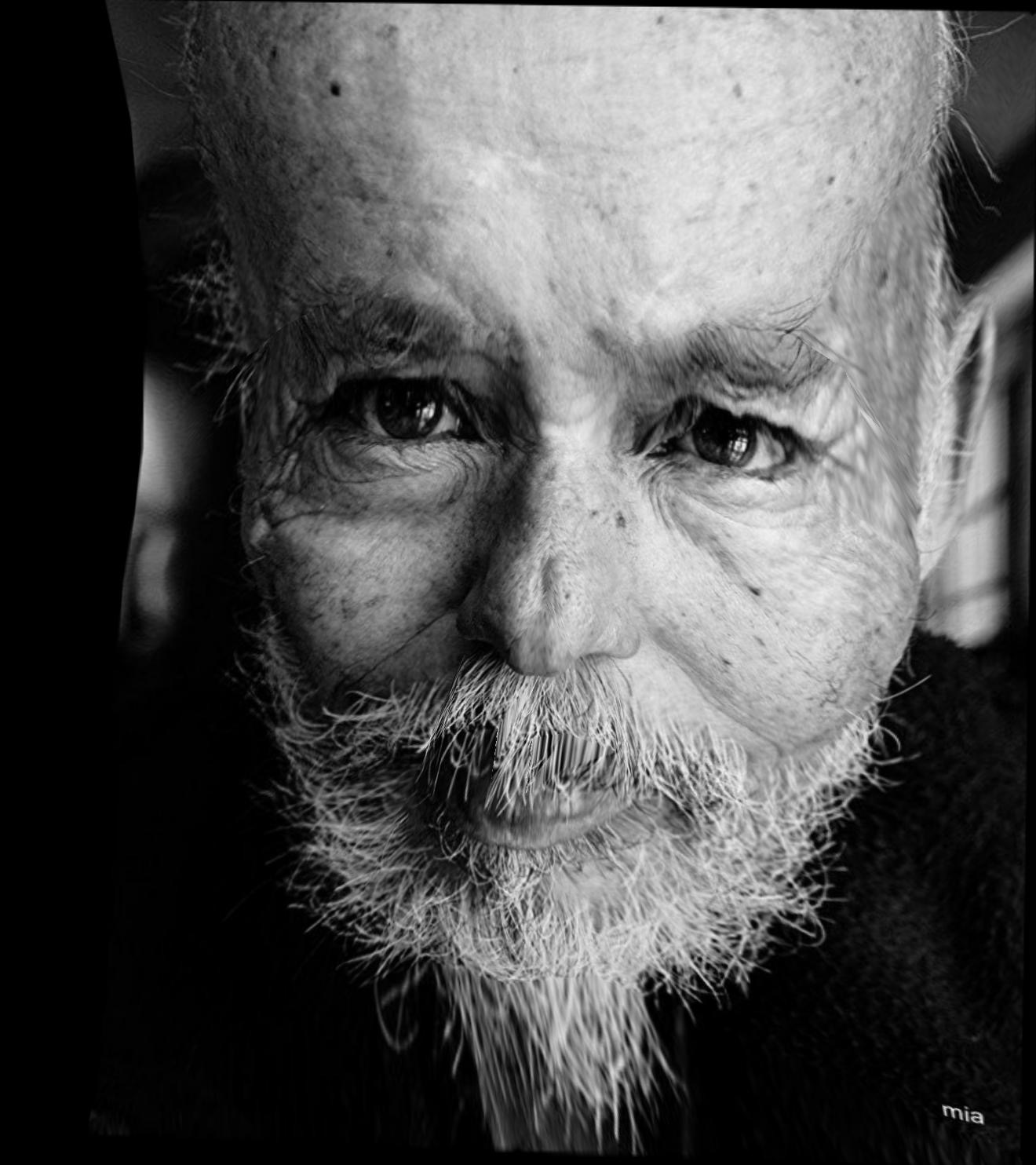} &
\includegraphics[width=0.875cm,height=0.973cm]{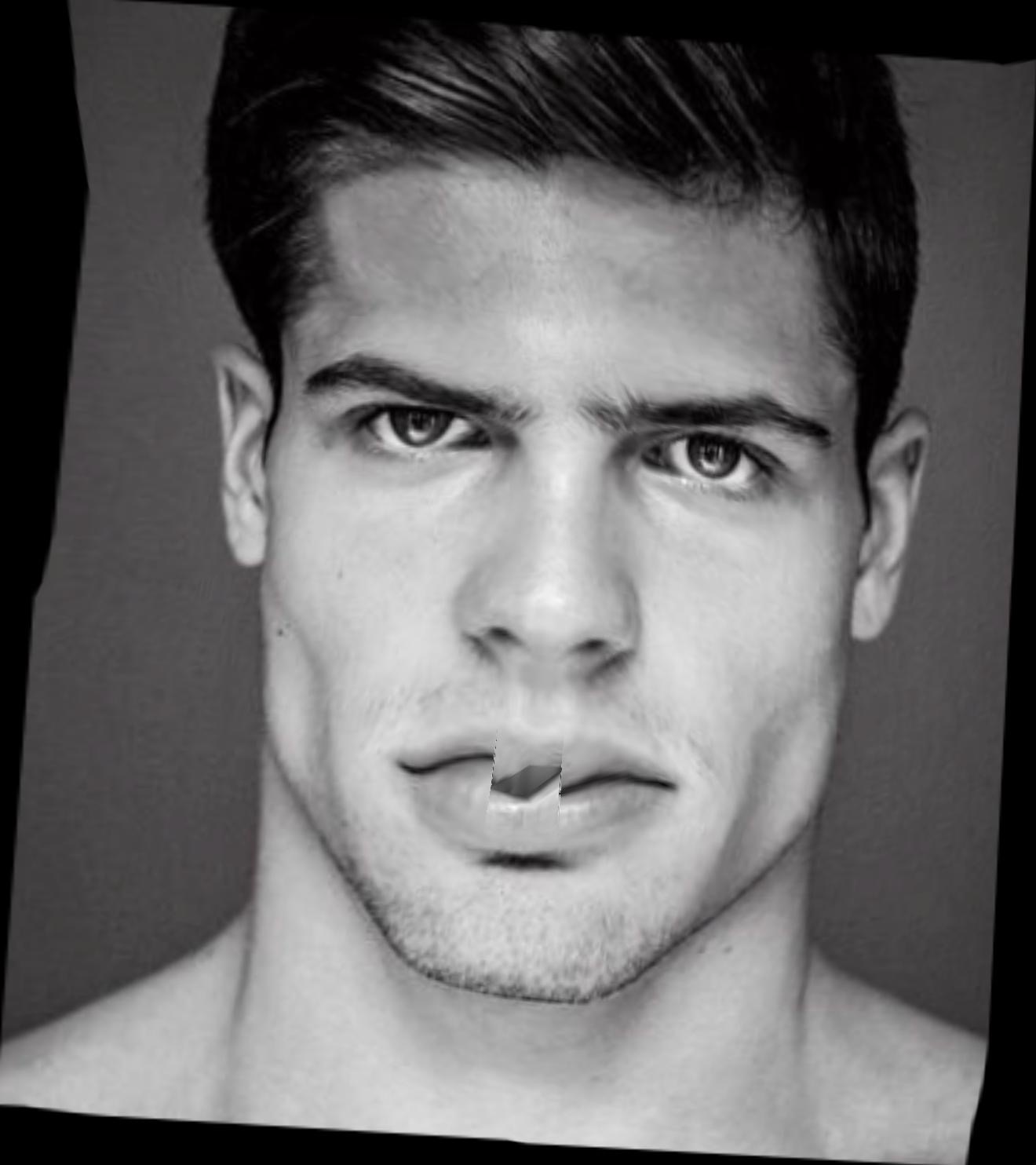} &
\includegraphics[width=0.864cm,height=0.968cm]{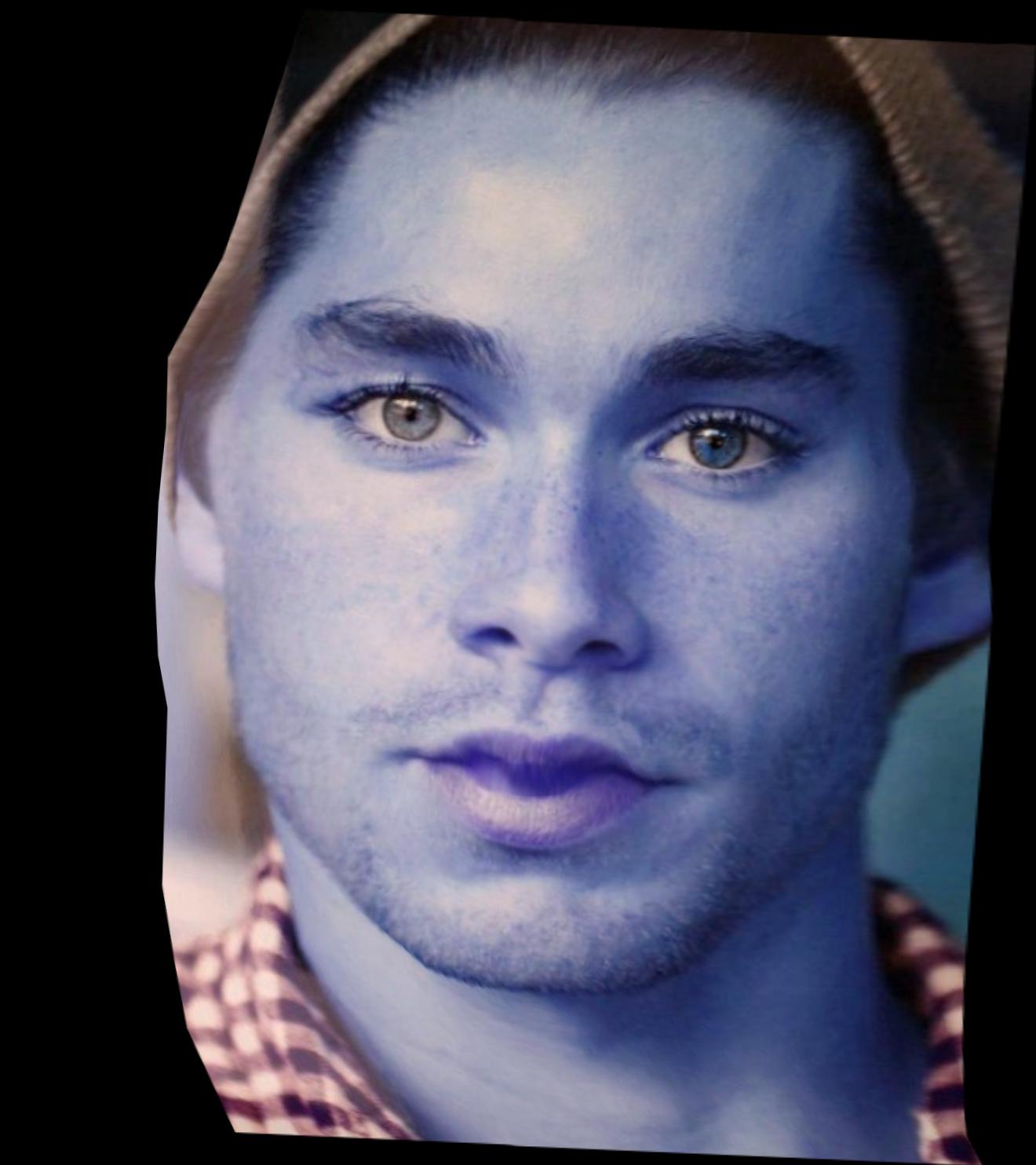} &
\includegraphics[width=0.829cm,height=0.943cm]{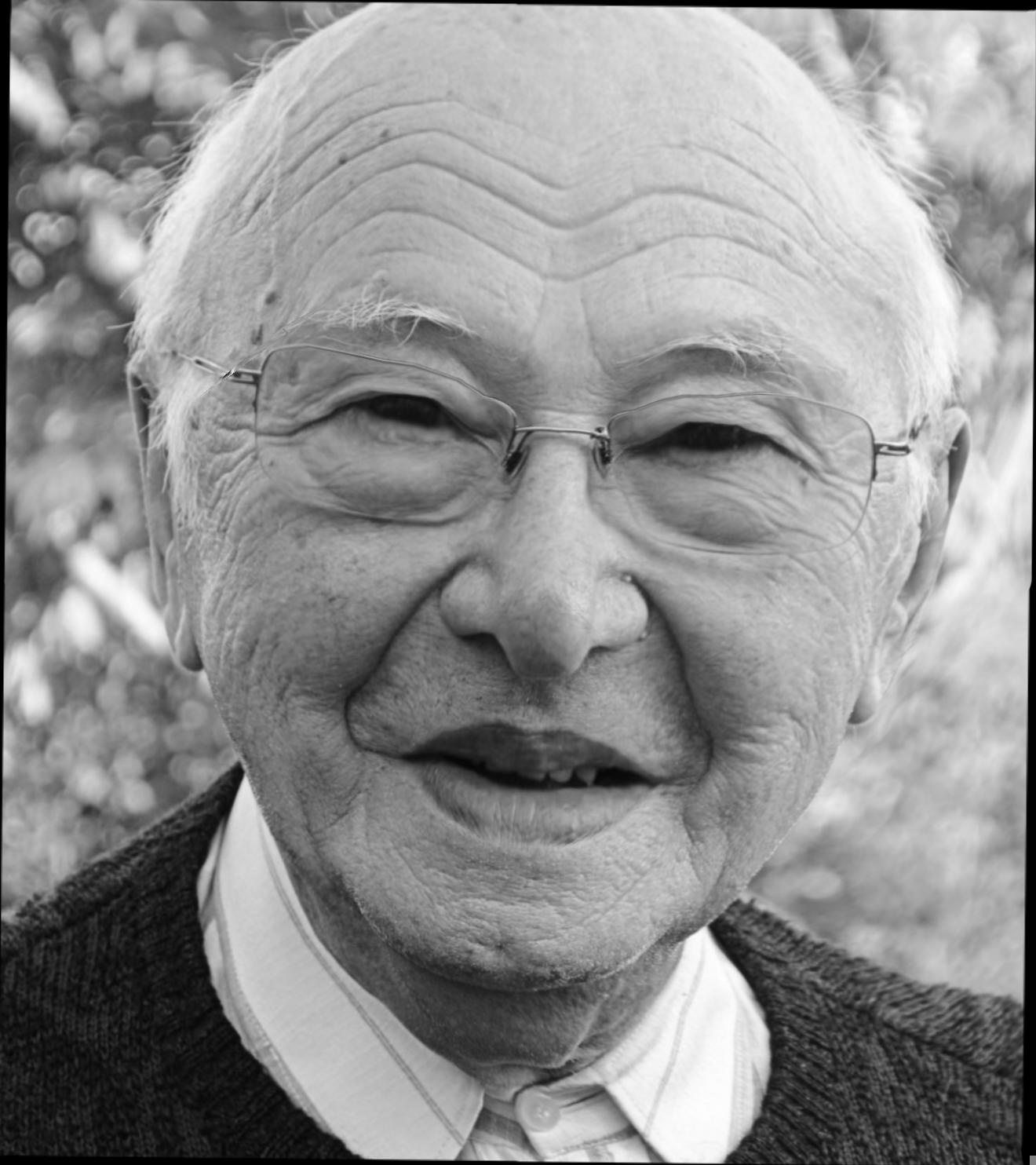} &
\includegraphics[width=0.851cm,height=0.942cm]{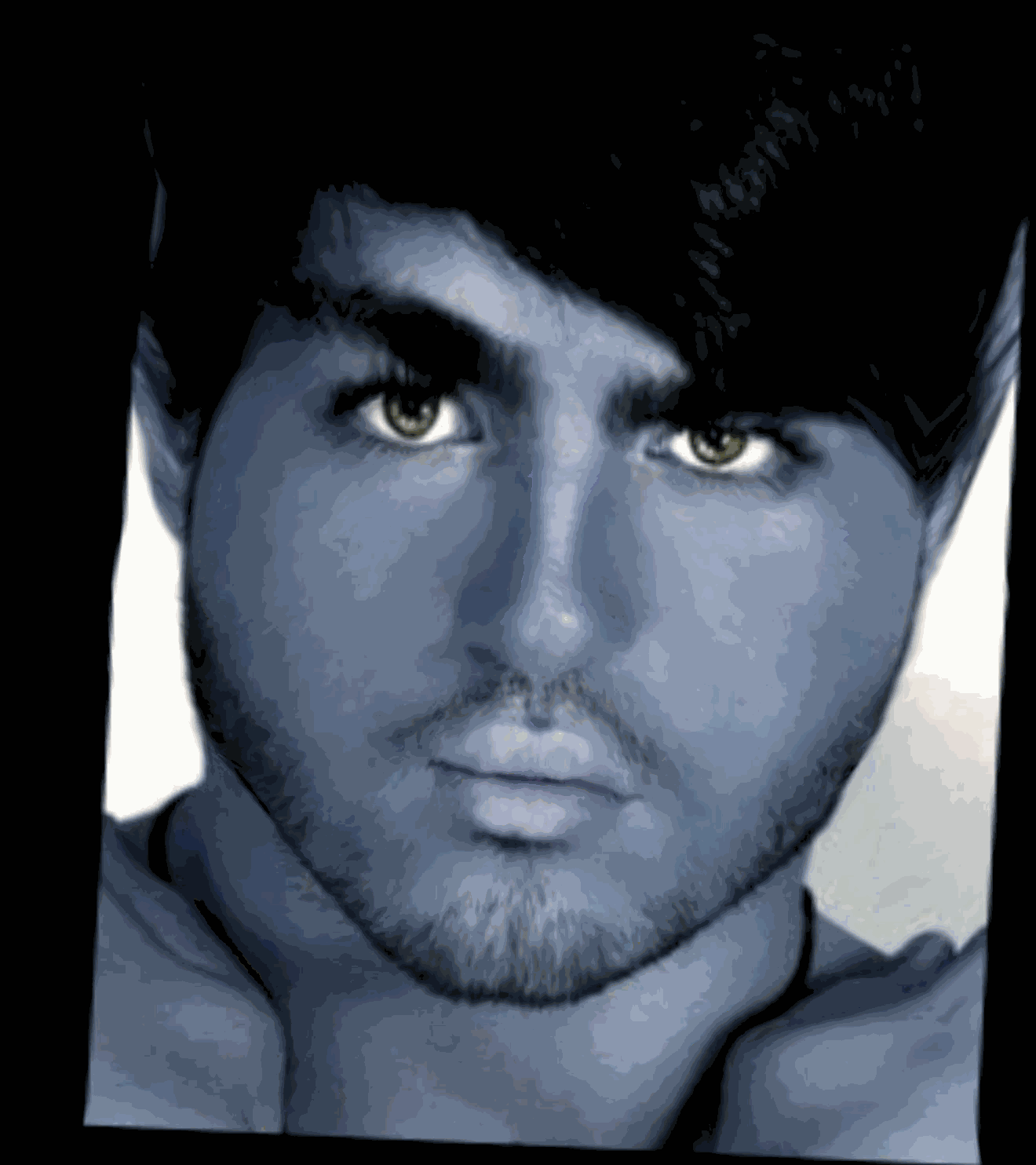} &
\includegraphics[width=0.811cm,height=0.942cm]{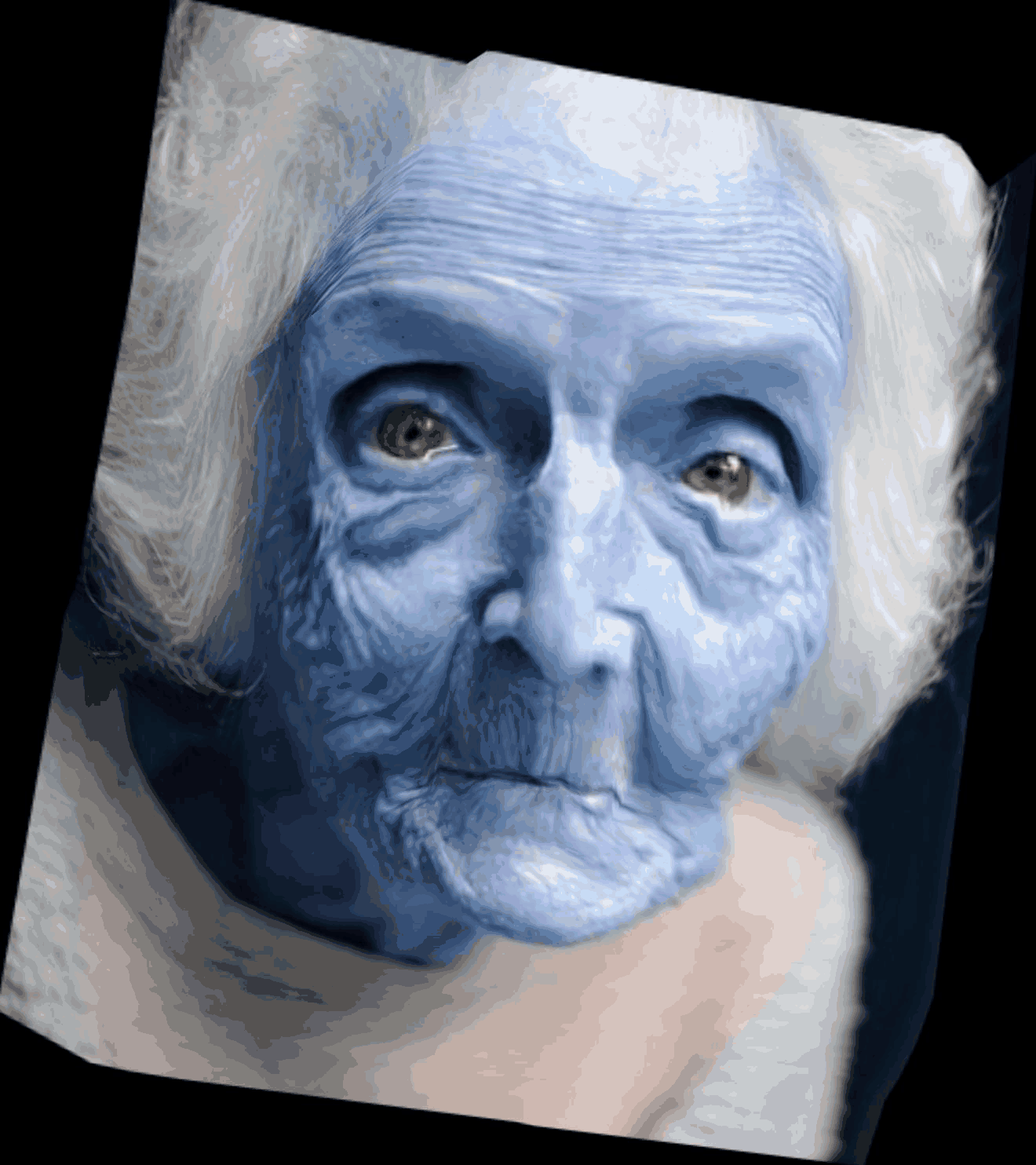} &
\includegraphics[width=0.890cm,height=1.007cm]{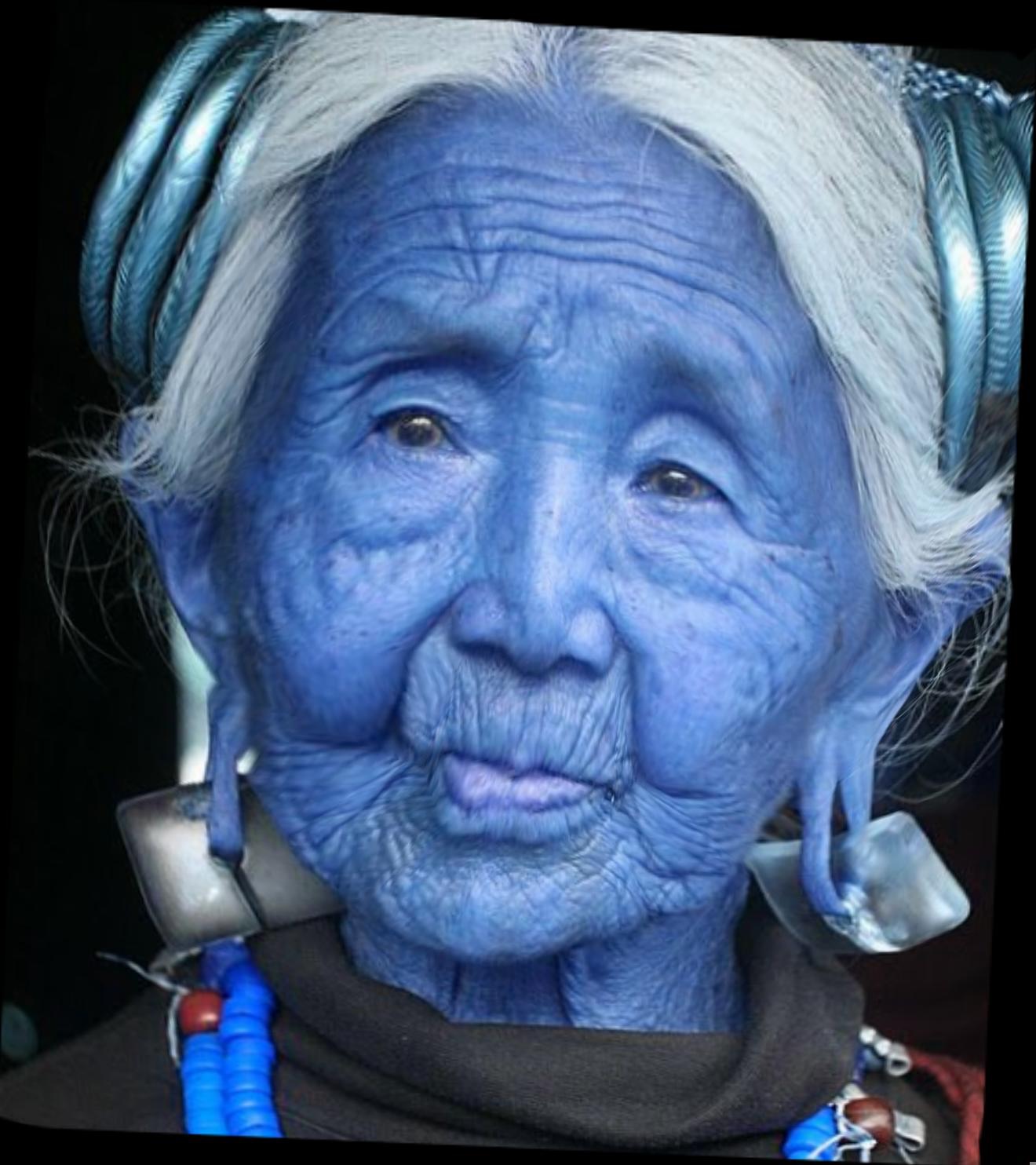} &
\includegraphics[width=0.847cm,height=0.942cm]{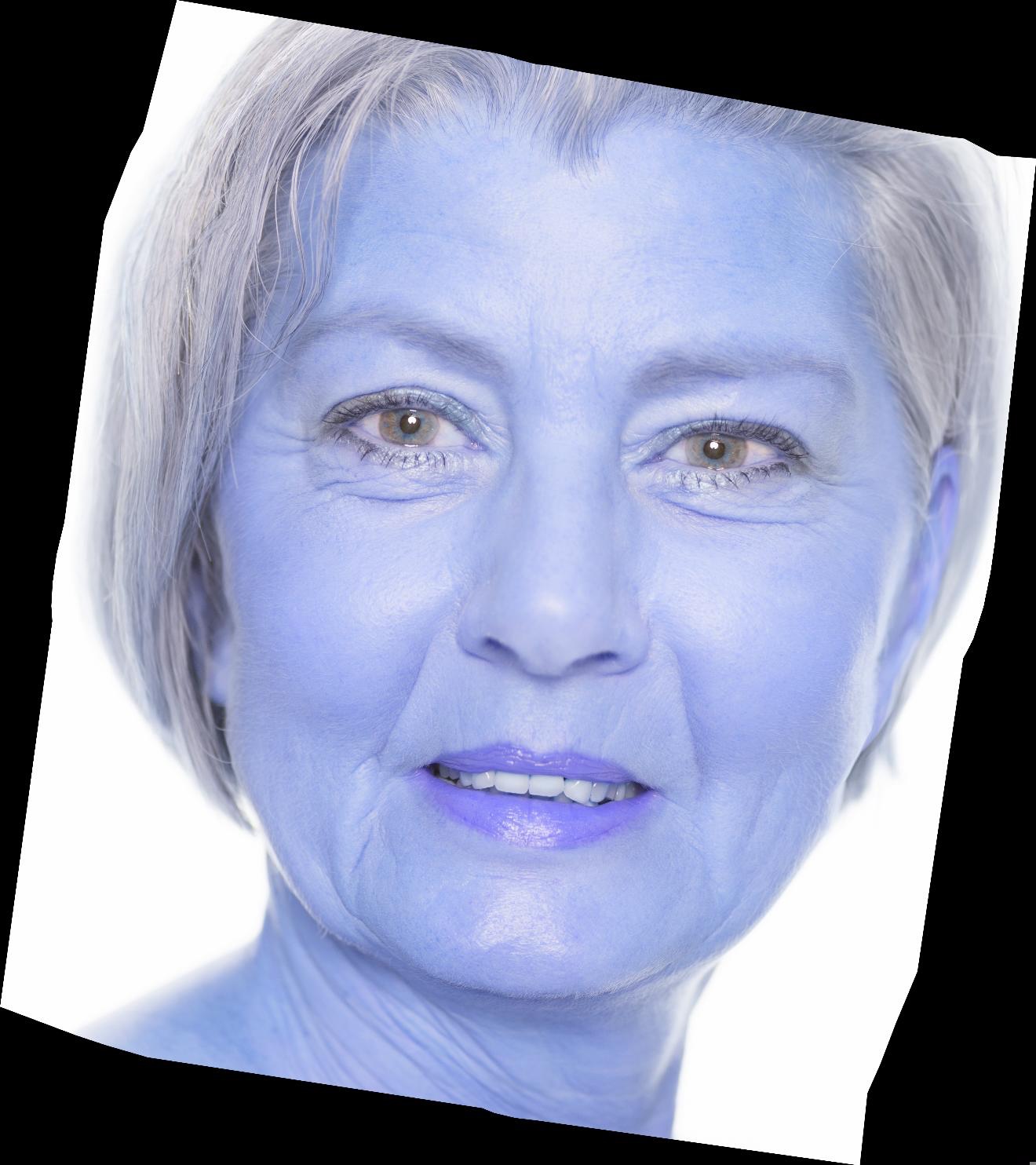} &
\includegraphics[width=0.893cm,height=0.968cm]{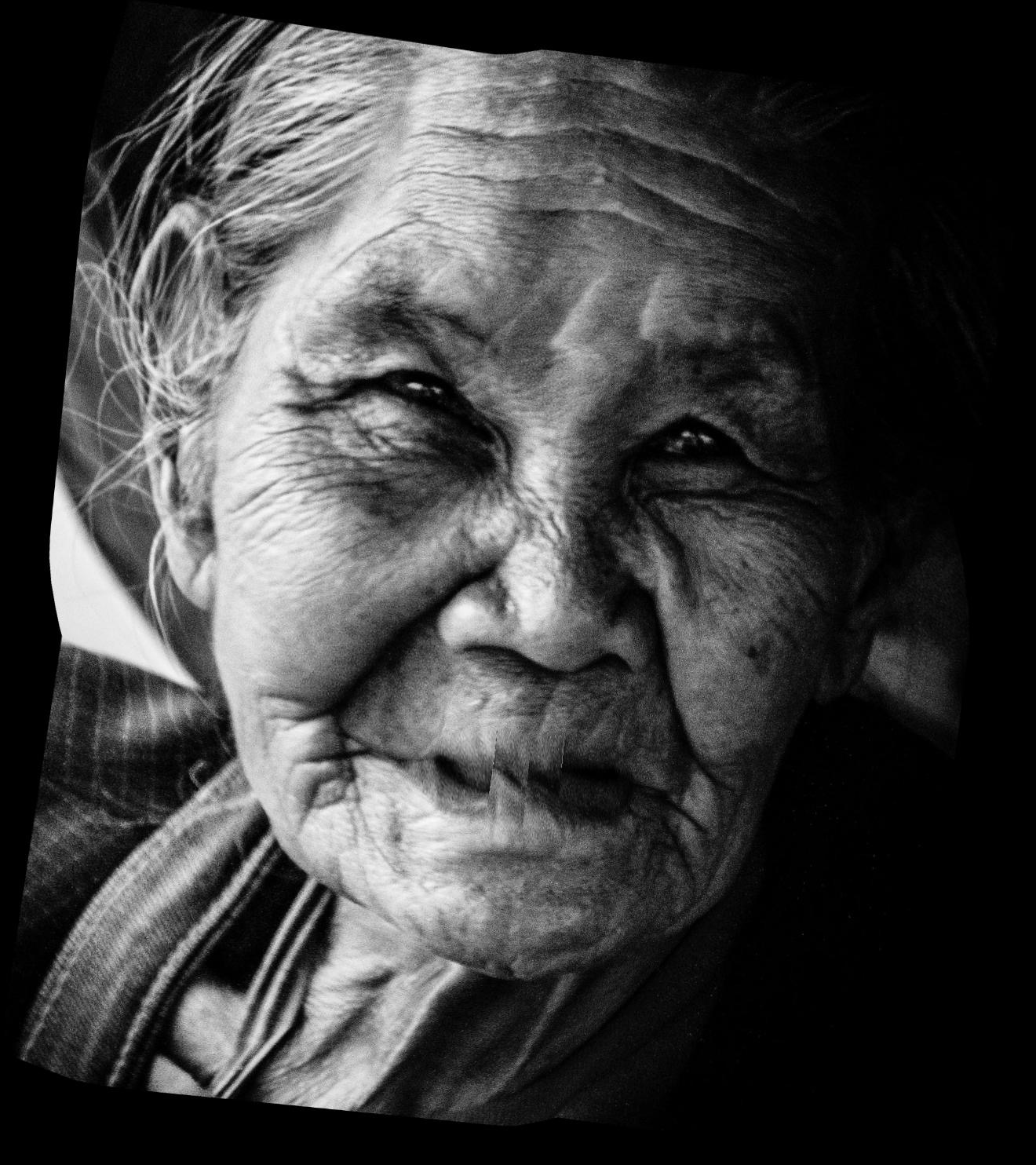} &
\includegraphics[width=0.826cm,height=0.942cm]{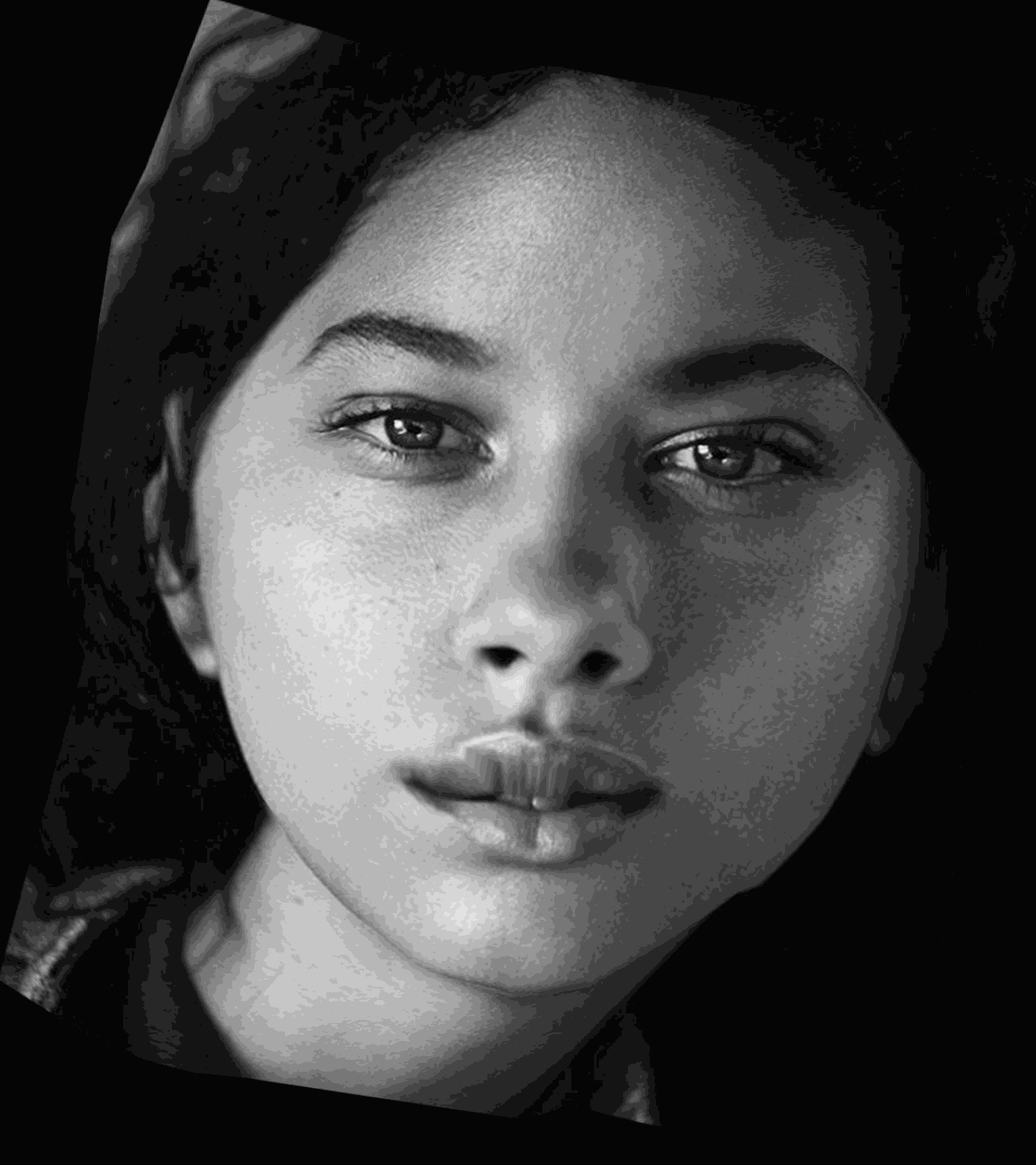} \\
\hline

\end{tabular}}
\end{table*}

\subsection{Unmixing Mixed Images Using the Second Independence Criterion}

Mixed images (right) are generated from source images (left); the proposed ICA variant should reconstruct the sources.

\begin{table}[htbp]
\caption{Classification accuracy on speech imagery (two classes). Baseline: FastICA Gauss. `$\wedge$' denotes $p<0.05$ (paired $t$-test).}
\label{tab:table3}
\resizebox{\columnwidth}{!}{%
\begin{tabular}{|l|c|c|c|c|c|}
\hline
\textbf{Setting} & \textbf{FI-Gauss} & \textbf{FI-tanh} & \textbf{FI-skew} & \textbf{Picard} & \textbf{Proposed} \\
\hline
W/O reg. & 53.3 & $53.3{\pm}0.0^{\wedge}$ & $53.3{\pm}0.0^{\wedge}$ & $60.0{\pm}2.0^{\wedge}$ & $\mathbf{76.4{\pm}3.5^{\wedge}}$ \\
\hline
With reg. & 50.0 & $50.0{\pm}0.0^{\wedge}$ & $66.7{\pm}3.2^{\wedge}$ & $66.7{\pm}3.8^{\wedge}$ & $\mathbf{79.8{\pm}2.9^{\wedge}}$ \\
\hline
Amari idx & 4.4 & 4.9 & 5.8 & 5.2 & \textbf{1.5} \\
\hline
\end{tabular}}
\end{table}

\subsection{Source Localization Using the Second Independence Criterion}

As shown in Table~\ref{tab:table3}, the proposed method achieves 76.4\% without regularization and 79.8\% with regularization---gains of 16.4\% and 13.1\% over PICARD, respectively---and a substantially lower Amari source separation index (1.5 vs.\ 4.4--5.8), confirming superior independent source extraction capability.

These results directly address RQ1: the direct probability product rule criterion outperforms all ICA baselines without requiring a fixed nonlinearity assumption.

\subsection{Results of Linear Models}

In Table~\ref{tab:table2}, Gender dataset eigenfaces are shown for different random initializations. Eigenface diversity under nullspace decorrelation is higher than under PCA. Eigenfaces in PCA appear more coarse-grained because eigenvectors are globally orthogonal, suppressing fine local structure.

Key findings from comparing eigenimages across methods:
\begin{itemize}
  \item Algorithm~\ref{pseudo:1} produces more diverse gray-level groups, indicating more independent components.
  \item Algorithm~\ref{pseudo:1} highlights more salient face subparts and digit variations.
  \item The diversity of eigenimages under Algorithm~\ref{pseudo:1} suggests implicit data augmentation (rotations and variations appear within single components).
  \item Algorithm~\ref{pseudo:1} better separates face subparts than PCA.
  \item Eigenfaces from nullspace decorrelation DR are less blurred and less constrained by orthonormality than in PCA.
\end{itemize}

As shown in Tables~\ref{tab:table6} and~\ref{tab:table7}, for linear models, eigenimage diversity, class representability, and subpart awareness are higher in WDIWCD than in WDDWCC, consistent with the greater flexibility of dependence over correlation.

These qualitative differences are quantified by the interpretability score (Manual Gender/Number Recognition Score) tabulated in Tables~\ref{tab:table6} and~\ref{tab:table7}. Regularized LDA achieves a score of only 3.5 on MNIST (Table~\ref{tab:table7}), compared with 5.8 for WDIWCD---a 66\% improvement---and 7.6 for Neural WDIWCD, confirming that the shift from correlation to dependence substantially enhances class-specific interpretability.

\subsection{Results of Neural Supervised Models}

Table~\ref{tab:table5} and part of Tables~\ref{tab:table6} ,~\ref{tab:table7} shows eigenimages of neural supervised models. Features are flattened and passed through a neural network before forming projection and nullspace features, optimized by either Algorithm~\ref{pseudo:1} (within-class correlation) or Algorithm~\ref{pseudo:2} (within-class dependence). Digit eigenimages are noticeably clearer in the dependence case than in the correlation case, while highly nonlinear subparts of faces are more readily extracted as eigenfaces in the decorrelation case. Patterns under Algorithm~\ref{pseudo:1} are also smoother and more natural, further confirming the complementary strengths of correlation and dependence objectives in neural settings.

\begin{table*}[!ht]
\caption{Eigenimages of neural supervised models. NWDDWCC = Neural Within-Class Correlation Whole-Data Decorrelation; NWDIWCD = Neural Whole-Data Independence Within-Class Dependence. The dependence-based method yields clearer and more salient digit representations.}
\centering
\label{tab:table5}
\renewcommand{\arraystretch}{1.5}
\resizebox{\textwidth}{!}{%
\begin{tabular}{|C{1.6cm}|C{1.6cm}|C{1.6cm}|C{1.6cm}|C{1.6cm}|C{1.6cm}|C{1.6cm}|C{1.6cm}|C{1.6cm}|C{1.6cm}|C{1.6cm}|}
\hline
\textbf{MNIST} & \textbf{0} & \textbf{1} & \textbf{2} & \textbf{3} & \textbf{4} & \textbf{5} & \textbf{6} & \textbf{7} & \textbf{8} & \textbf{9} \\
\hline
NWCCWDD
  & \includegraphics[width=1.2cm,height=1.2cm]{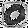}
  & \includegraphics[width=1.2cm,height=1.2cm]{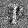}
  & \makecell{Not\\Recognize\\able}
  & \includegraphics[width=1.2cm,height=1.2cm]{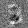}
  & \includegraphics[width=1.2cm,height=1.2cm]{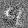}
  & \includegraphics[width=1.2cm,height=1.2cm]{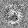}
  & \includegraphics[width=1.2cm,height=1.2cm]{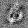}
  & \includegraphics[width=1.2cm,height=1.2cm]{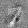}
  & \includegraphics[width=1.2cm,height=1.2cm]{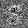}
  & \includegraphics[width=1.2cm,height=1.2cm]{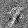} \\
\hline
NWDIWCD
  & \includegraphics[width=1.2cm,height=1.2cm]{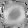}
  & \includegraphics[width=1.2cm,height=1.2cm]{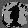}
  & \makecell{Not\\Recognize\\able}
  & \includegraphics[width=1.2cm,height=1.2cm]{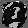}
  & \includegraphics[width=1.2cm,height=1.2cm]{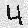}
  & \includegraphics[width=1.2cm,height=1.2cm]{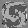}
  & \includegraphics[width=1.2cm,height=1.2cm]{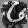}
  & \includegraphics[width=1.2cm,height=1.2cm]{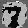}
  & \makecell{Not\\Recognize\\able}
  & \includegraphics[width=1.2cm,height=1.2cm]{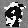} \\
\hline
\end{tabular}}
\end{table*}

\begin{figure}[htbp]
  \centering
  \includegraphics[width=\columnwidth]{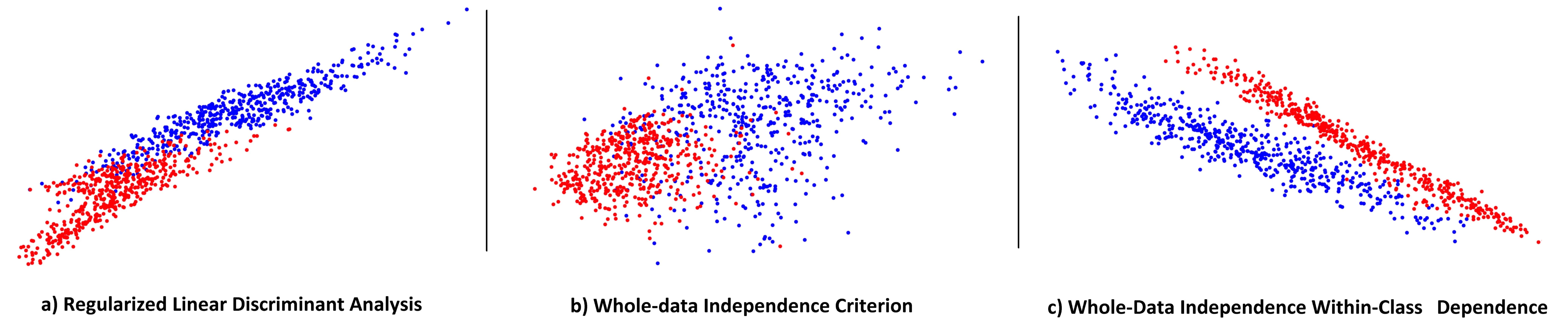}
  \caption{Class separability of each method compared to WDIWCD.}
  \label{fig:classsep}
\end{figure}

\begin{figure}[htbp]
  \centering
  \includegraphics[width=\columnwidth]{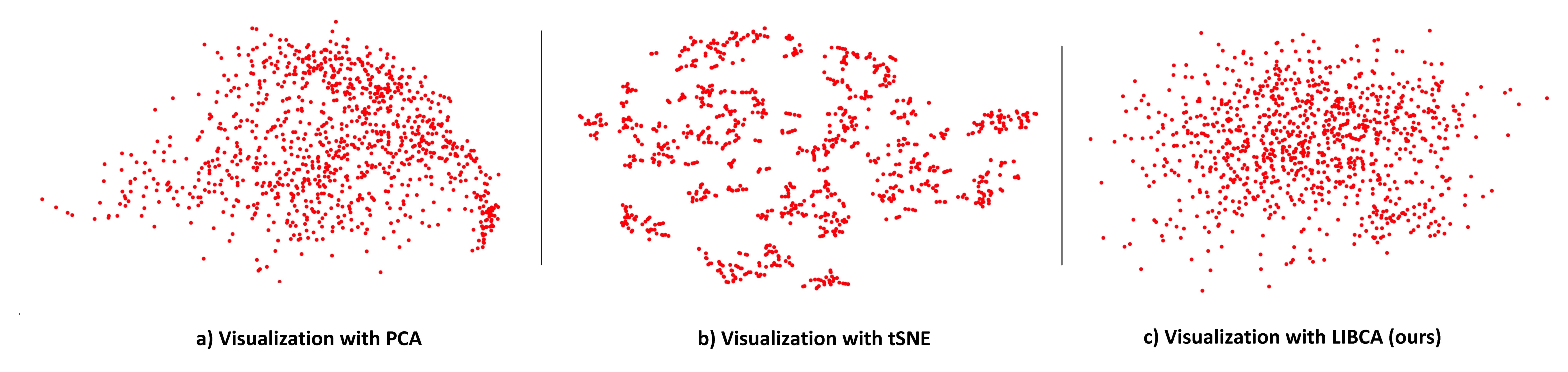}
  \caption{Contrast power of conventional vs.\ proposed dependence-based approach. Each point represents a 2D projection. The proposed LIBCA algorithm shows higher contrast and spread than t-SNE and PCA.}
  \label{fig:contrast}
\end{figure}

\subsection{Results of Layer Sharing with VAE}

The first VAE layer is shared with NWDIWCD, using the combined loss in~(\ref{eq:28}). Table~\ref{tab:table6} presents the obtained class-specific eigenimage comparisons across all neural layer-sharing models for the Gender dataset, and Table~\ref{tab:table7} presents the corresponding results for MNIST. The last layer produces smoother eigenimages than the first layer because WDIWCD reinforces small discriminative patterns in the first layer, enabling subsequent layers to focus on coarser global structure. Patterns present in one target class but absent in others are the most informative for interpretable machine learning.

\textbf{Interpretability score analysis (RQ2 and RQ3).}
Comparing WDIWCD (score 5.8) to WDDWCC (score 4.8) in Table~\ref{tab:table6} shows a 20.8\% interpretability gain, confirming that generalizing from correlation to dependence yields measurably higher interpretability (RQ2). In Table~\ref{tab:table7}, Neural WDIWCD achieves 7.6 vs.\ Neural WDDWCC at 5.9 (a 28.8\% gap), further validating dependence-based objectives. VAE-combined methods consistently outperform their individual counterparts: VAE-WDIWCD (last layer) achieves 4.5 on MNIST and 7.1 on Gender, substantially higher than standalone WDIWCD (5.8 Gender, 2.7 MNIST null-space decorrelation equivalent), supporting RQ3.

\begin{table*}[!ht]
\caption{Comparison of all methods layer-shared with a Variational Autoencoder for the Gender dataset. MGRS = Manual Gender Recognition Score (interpretability). WDDWCC = Whole-Data Decorrelation Within-Class Correlation; WDIWCD = Whole-Data Independence Within-Class Dependence; NWDIWCD = Neural WDIWCD; CECL = Class Entities Largest Correlation; MWCCE = Max Within-Class Cross-Entropy. Numerical scores are means over five independent runs; `$\wedge$' denotes statistical significance ($p<0.05$, paired $t$-test).}
\centering
\label{tab:table6}
\scriptsize
\setlength{\tabcolsep}{2pt}
\resizebox{\textwidth}{!}{%
\begin{tabular}{|l|*{10}{c|}r|}
\hline
\textbf{Weight parameter} &
\multicolumn{5}{c|}{$\longleftarrow$\textemdash\textemdash\textemdash\textemdash\textemdash\textemdash male\textemdash\textemdash\textemdash\textemdash\textemdash\textemdash$\longrightarrow$} &
\multicolumn{5}{c|}{$\longleftarrow$\textemdash\textemdash\textemdash\textemdash\textemdash\textemdash female\textemdash\textemdash\textemdash\textemdash\textemdash\textemdash$\longrightarrow$} &
\textbf{MGRS} \\
\hline

Dataset Sample &
\includegraphics[width=1.5cm,height=1.5cm]{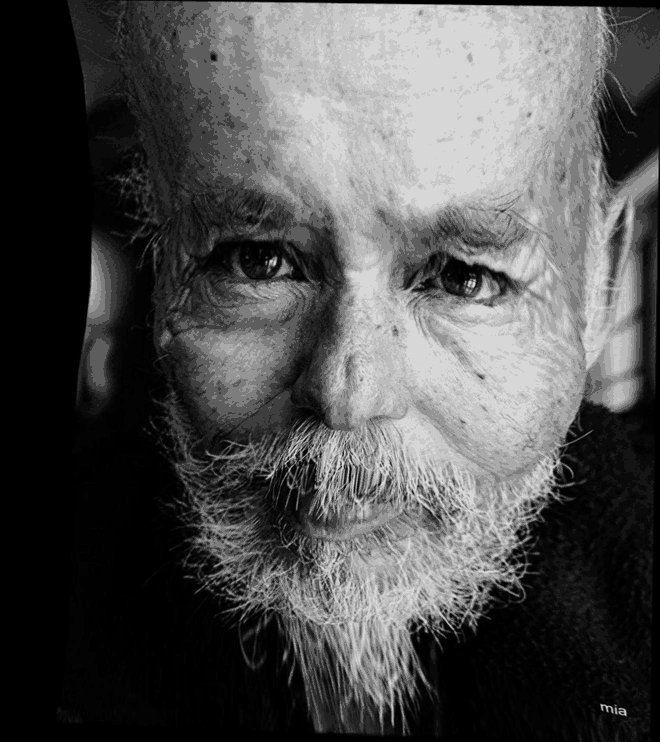} &
\includegraphics[width=1.5cm,height=1.5cm]{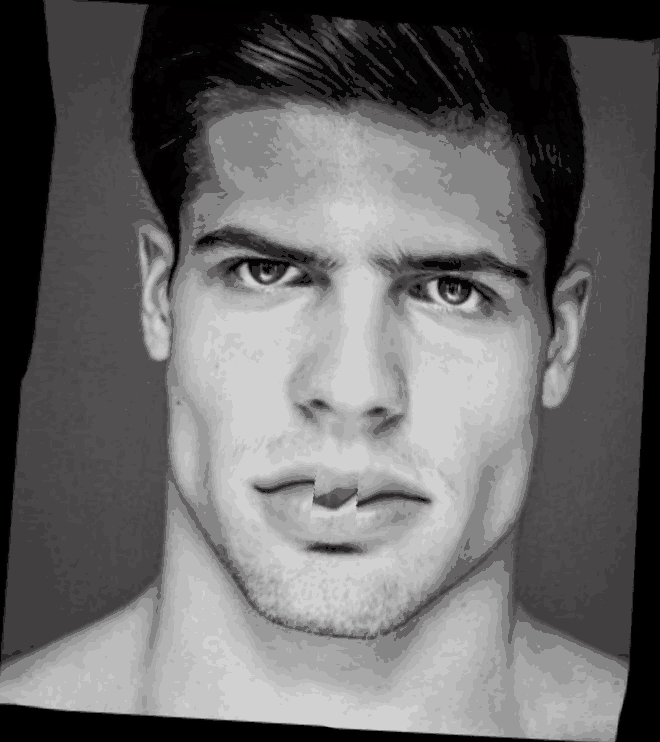} &
\includegraphics[width=1.5cm,height=1.5cm]{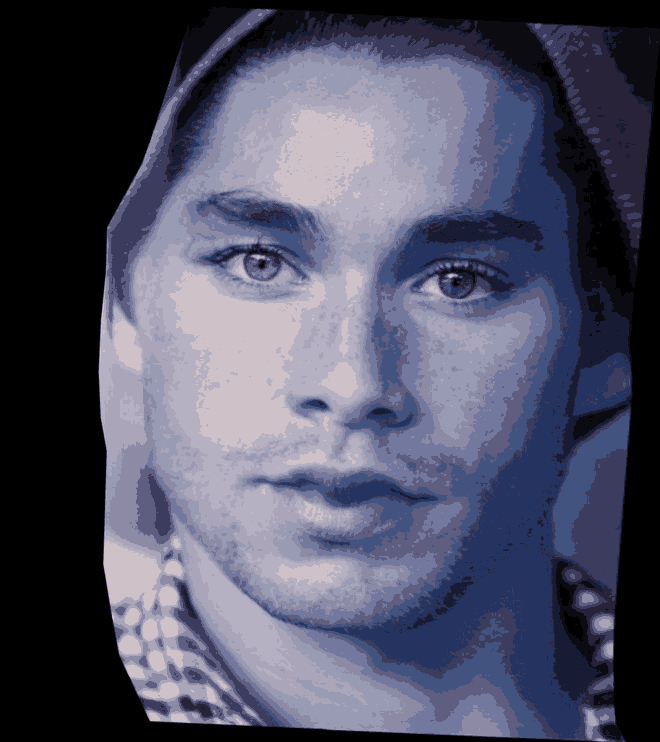} &
\includegraphics[width=1.5cm,height=1.5cm]{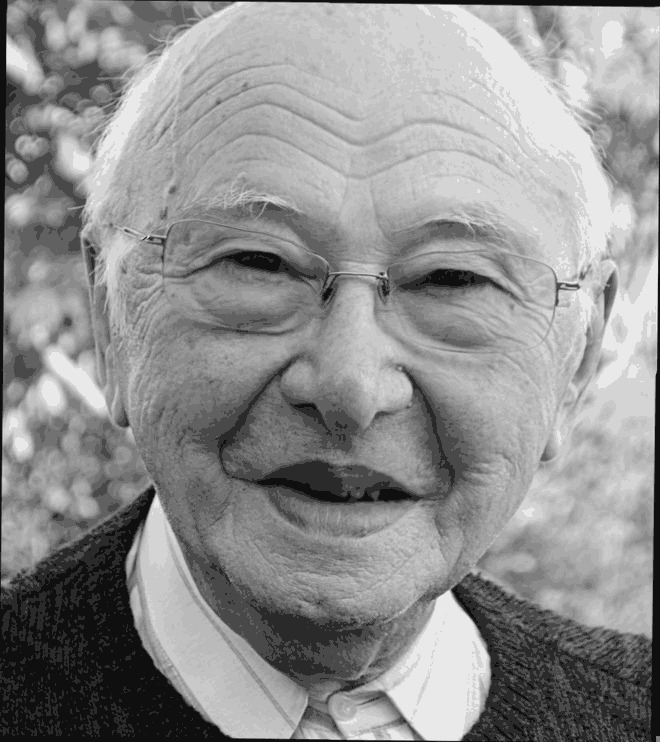} &
\includegraphics[width=1.5cm,height=1.5cm]{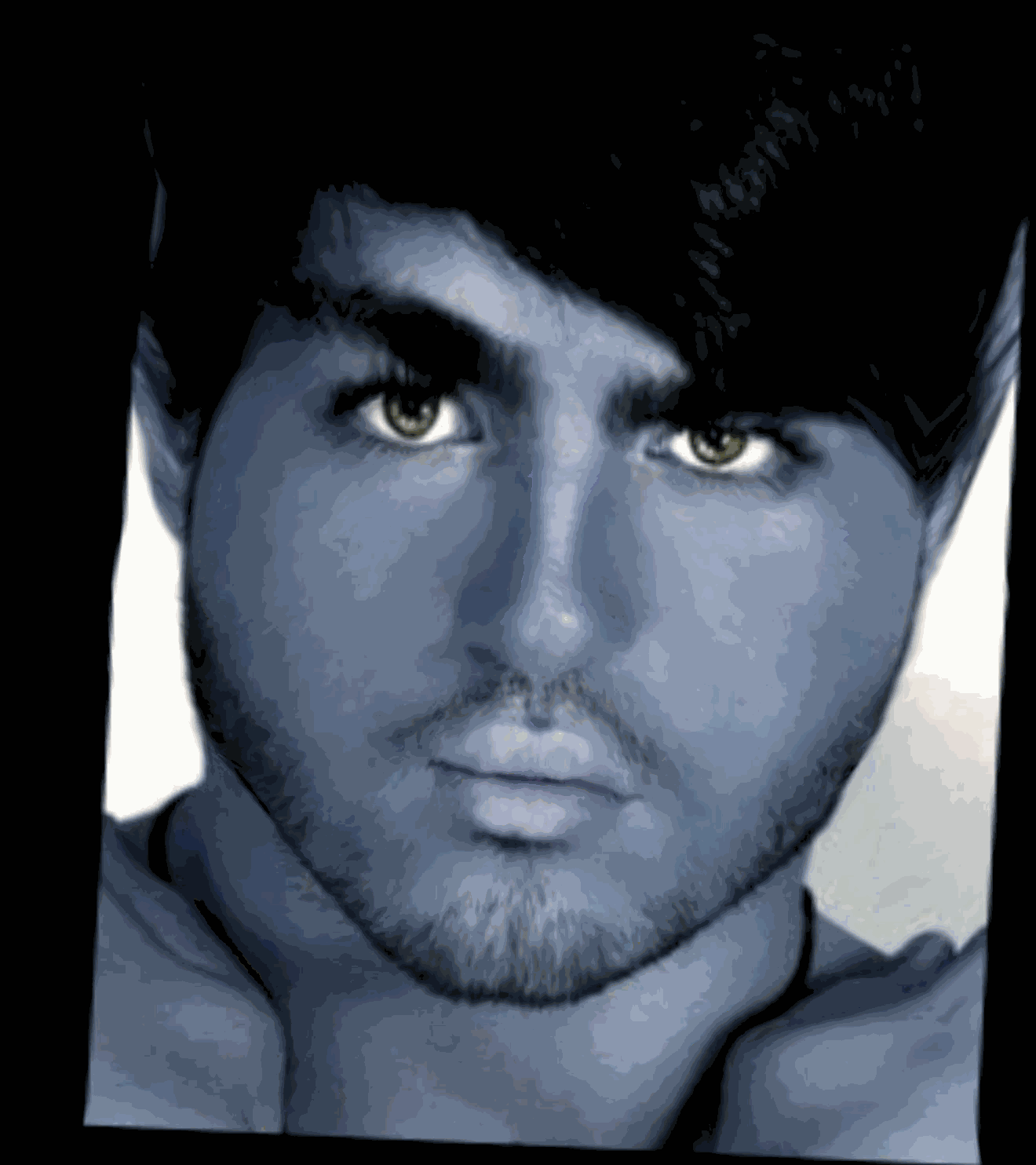} &
\includegraphics[width=1.5cm,height=1.5cm]{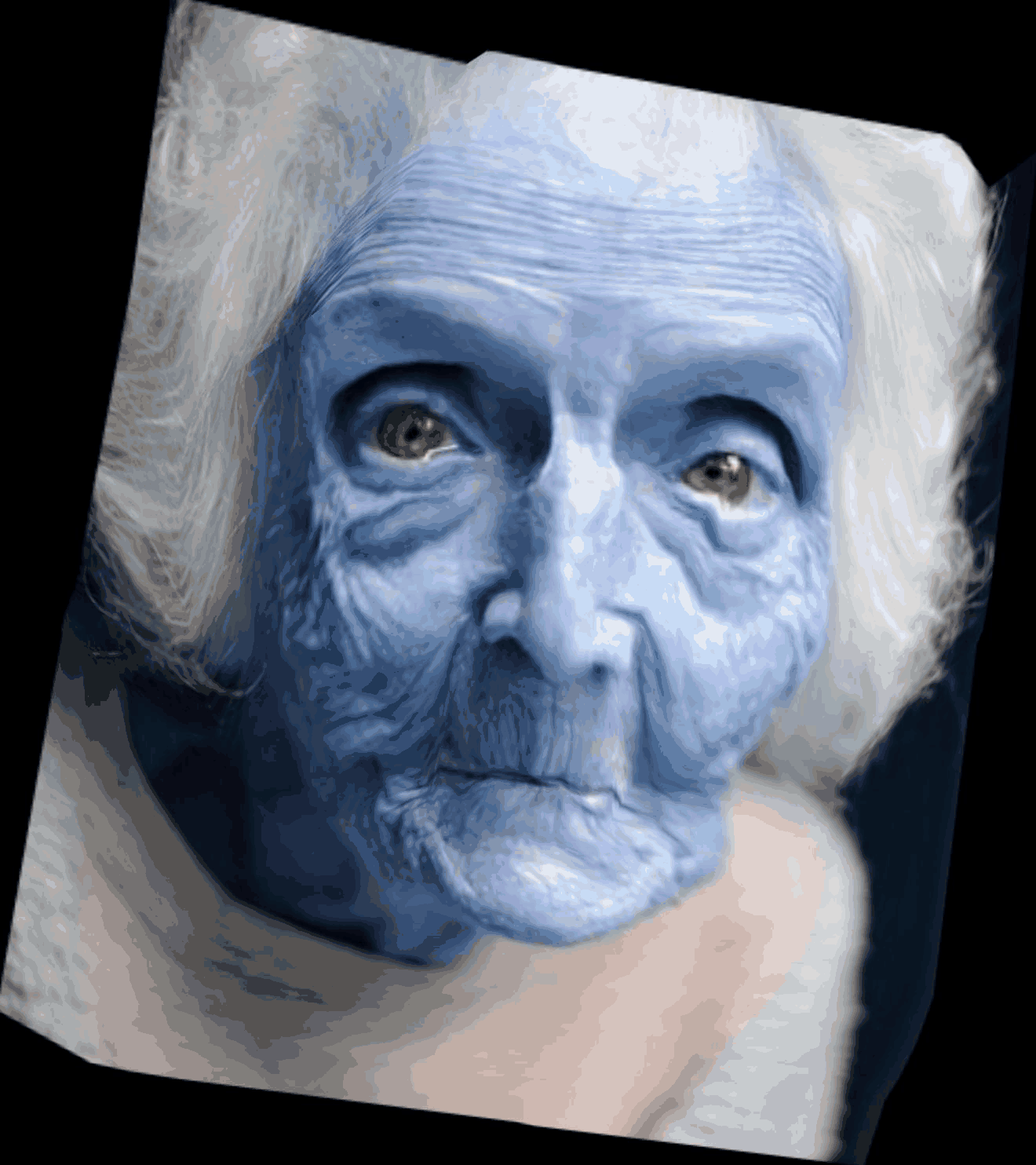} &
\includegraphics[width=1.5cm,height=1.5cm]{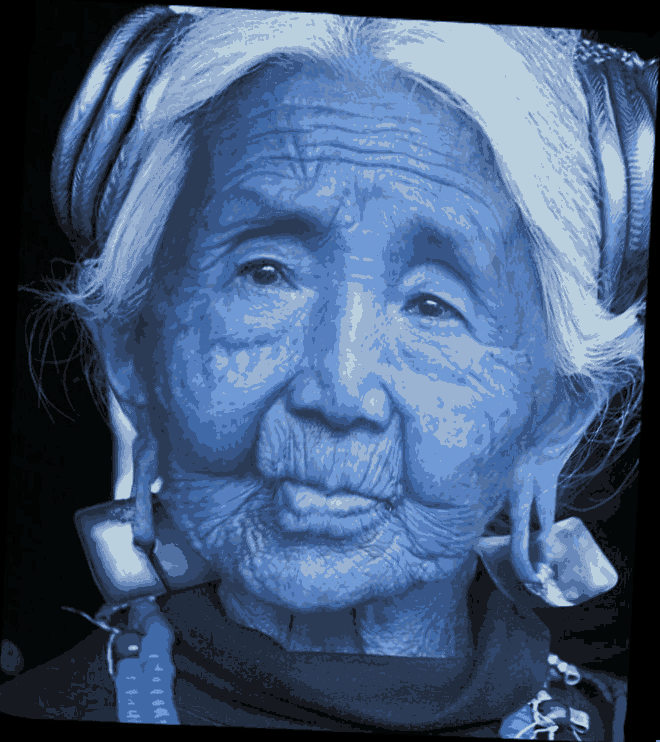} &
\includegraphics[width=1.5cm,height=1.5cm]{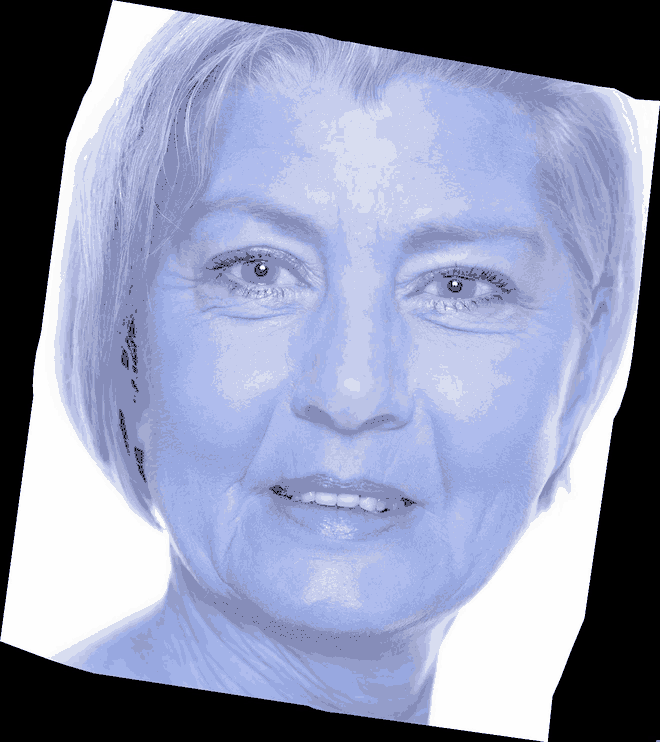} &
\includegraphics[width=1.5cm,height=1.5cm]{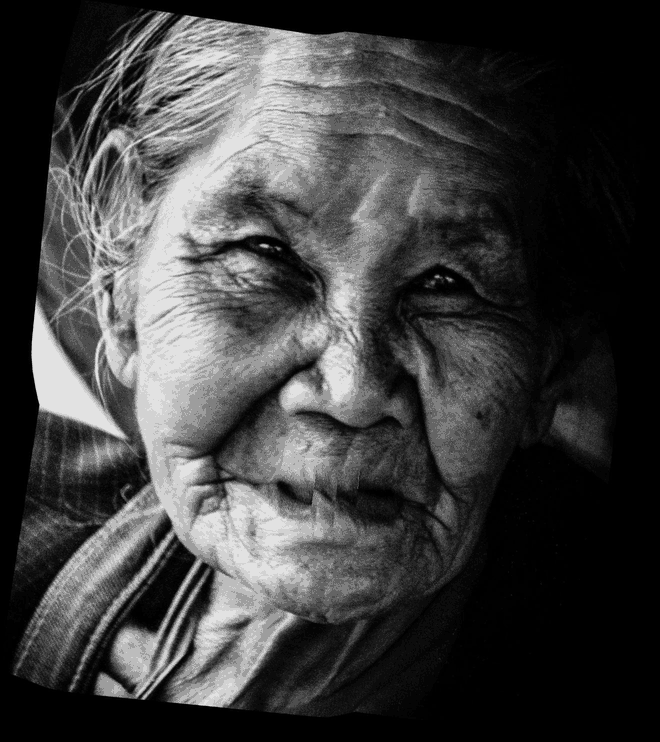} &
\includegraphics[width=1.5cm,height=1.5cm]{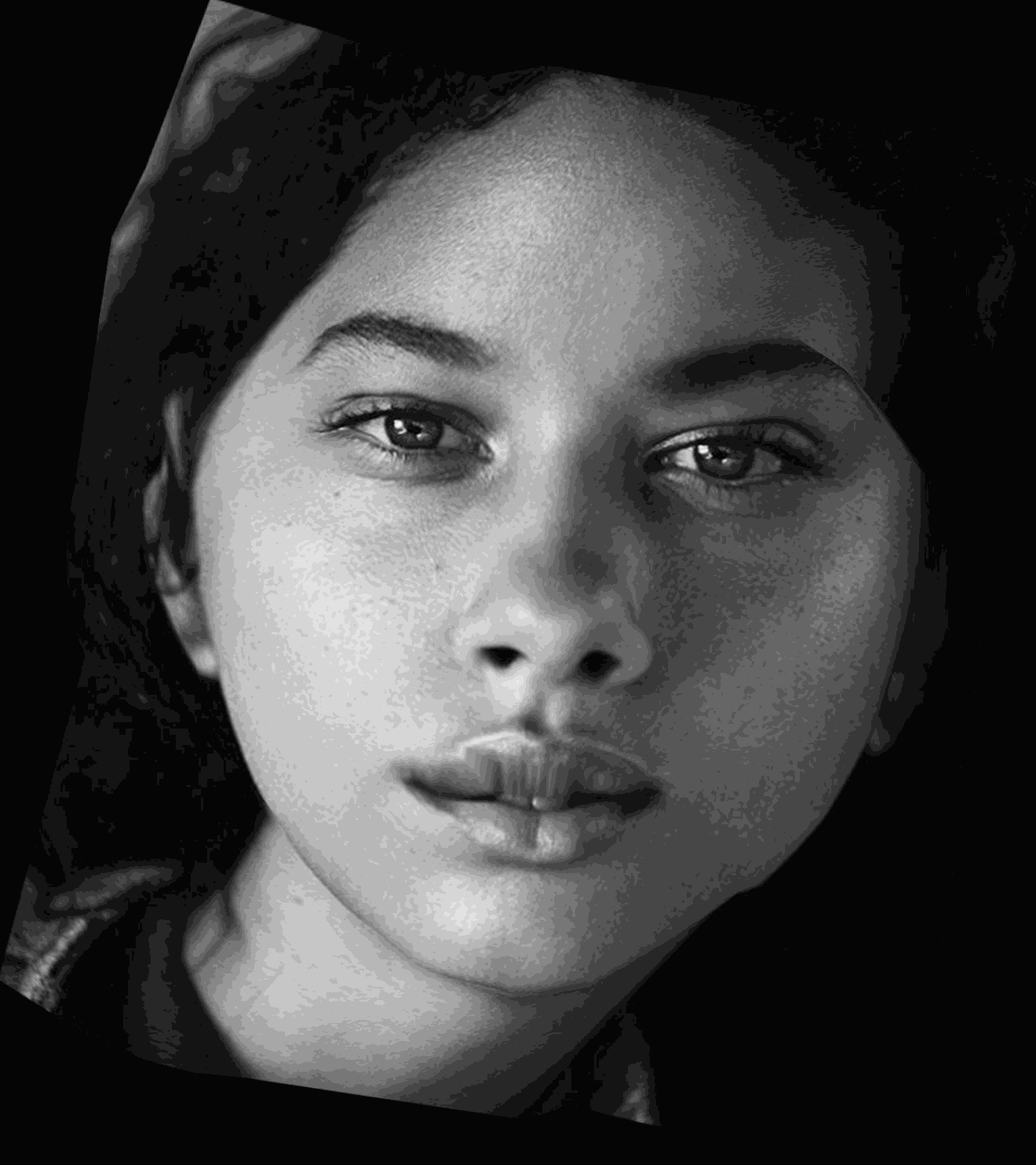} &
\makecell{$1{+}1{+}1{+}1{+}1{+}1{+}1{+}1{+}1{+}1{=}$ \\ \textbf{10}} \\
\hline

\makecell[l]{VAE-WDDWCC Last \\ Layer \textbf{(proposed)}} &
\includegraphics[width=1.5cm,height=1.5cm]{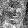} &
\includegraphics[width=1.5cm,height=1.5cm]{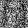} &
\includegraphics[width=1.5cm,height=1.5cm]{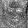} &
\includegraphics[width=1.5cm,height=1.5cm]{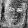} &
\includegraphics[width=1.5cm,height=1.5cm]{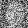} &
\includegraphics[width=1.5cm,height=1.5cm]{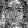} &
\includegraphics[width=1.5cm,height=1.5cm]{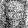} &
\includegraphics[width=1.5cm,height=1.5cm]{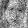} &
\includegraphics[width=1.5cm,height=1.5cm]{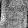} &
\includegraphics[width=1.5cm,height=1.5cm]{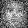} &
\makecell{$0.6{+}0.5{+}1{+}1{+}0.5{+}1{+}0.7{+}1{+}0.9{+}1{=}$ \\ \textbf{8.2}} \\
\hline

\makecell[l]{NWDIWCD \\ \textbf{(proposed)}} &
\includegraphics[width=1.5cm,height=1.5cm]{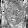} &
\includegraphics[width=1.5cm,height=1.5cm]{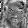} &
\includegraphics[width=1.5cm,height=1.5cm]{images/image21.png} &
\makecell{Not \\ Recognize\\ able} &
\includegraphics[width=1.5cm,height=1.5cm]{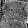} &
\includegraphics[width=1.5cm,height=1.5cm]{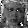} &
\includegraphics[width=1.5cm,height=1.5cm]{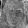} &
\includegraphics[width=1.5cm,height=1.5cm]{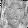} &
\includegraphics[width=1.5cm,height=1.5cm]{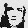} &
\includegraphics[width=1.5cm,height=1.5cm]{images/image25.png} &
\makecell{$0.6{+}1{+}0.6{+}0{+}1{+}1{+}0.8{+}0.9{+}1{+}0.8{=}$ \\ \textbf{7.7}} \\
\hline

\makecell[l]{VAE-WDIWCD First layer \\ \textbf{(proposed)}} &
\includegraphics[width=1.5cm,height=1.5cm]{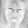} &
\includegraphics[width=1.5cm,height=1.5cm]{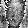} &
\includegraphics[width=1.5cm,height=1.5cm]{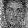} &
\includegraphics[width=1.5cm,height=1.5cm]{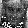} &
\includegraphics[width=1.5cm,height=1.5cm]{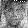} &
\includegraphics[width=1.5cm,height=1.5cm]{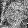} &
\includegraphics[width=1.5cm,height=1.5cm]{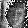} &
\includegraphics[width=1.5cm,height=1.5cm]{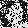} &
\includegraphics[width=1.5cm,height=1.5cm]{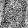} &
\includegraphics[width=1.5cm,height=1.5cm]{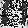} &
\makecell{$0.9{+}0.6{+}1{+}0.6{+}1{+}0.8{+}0.8{+}0.6{+}0.8{=}$ \\ \textbf{7.1}} \\
\hline

\makecell[l]{VAE-WDDWCC First \\ Layer \textbf{(proposed)}} &
\includegraphics[width=1.5cm,height=1.5cm]{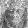} &
\includegraphics[width=1.5cm,height=1.5cm]{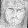} &
\includegraphics[width=1.5cm,height=1.5cm]{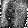} &
\includegraphics[width=1.5cm,height=1.5cm]{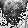} &
\includegraphics[width=1.5cm,height=1.5cm]{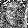} &
\includegraphics[width=1.5cm,height=1.5cm]{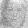} &
\includegraphics[width=1.5cm,height=1.5cm]{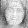} &
\includegraphics[width=1.5cm,height=1.5cm]{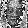} &
\makecell{Not \\ Recognize\\ able} &
\makecell{Not \\ Recognize\\ able} &
\makecell{$0.9{+}1{+}0.6{+}0.8{+}0.6{+}0.9{+}1{+}1{+}0{+}0{=}$ \\ \textbf{6.8}} \\
\hline

\makecell[l]{Null-space Decorrelation \\ (Unsup.)} &
\includegraphics[width=1.5cm,height=1.5cm]{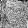} &
\includegraphics[width=1.5cm,height=1.5cm]{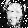} &
\includegraphics[width=1.5cm,height=1.5cm]{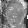} &
\includegraphics[width=1.5cm,height=1.5cm]{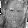} &
\includegraphics[width=1.5cm,height=1.5cm]{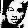} &
\includegraphics[width=1.5cm,height=1.5cm]{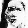} &
\includegraphics[width=1.5cm,height=1.5cm]{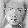} &
\includegraphics[width=1.5cm,height=1.5cm]{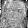} &
\includegraphics[width=1.5cm,height=1.5cm]{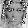} &
\includegraphics[width=1.5cm,height=1.5cm]{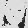} &
\makecell{$0.5{+}0.4{+}0.6{+}0.6{+}1{+}0.8{+}1{+}0.4{+}0.6{+}0.7{=}$ \\ \textbf{6.6}} \\
\hline

\makecell[l]{VAE Last Layer \\ \textbf{(baseline)}} &
\includegraphics[width=1.5cm,height=1.5cm]{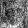} &
\includegraphics[width=1.5cm,height=1.5cm]{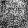} &
\includegraphics[width=1.5cm,height=1.5cm]{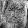} &
\includegraphics[width=1.5cm,height=1.5cm]{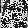} &
\includegraphics[width=1.5cm,height=1.5cm]{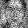} &
\includegraphics[width=1.5cm,height=1.5cm]{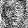} &
\includegraphics[width=1.5cm,height=1.5cm]{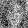} &
\includegraphics[width=1.5cm,height=1.5cm]{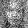} &
\includegraphics[width=1.5cm,height=1.5cm]{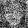} &
\makecell{Not \\ Recognize\\ able} &
\makecell{$0.8{+}0.8{+}0.8{+}0.6{+}0.4{+}0.5{+}0.9{+}0.9{+}0.8{+}0{=}$ \\ \textbf{6.5}} \\
\hline

\makecell[l]{VAE-WDIWCD Last Layer \\ \textbf{(proposed)}} &
\includegraphics[width=1.5cm,height=1.5cm]{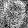} &
\includegraphics[width=1.5cm,height=1.5cm]{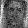} &
\includegraphics[width=1.5cm,height=1.5cm]{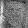} &
\includegraphics[width=1.5cm,height=1.5cm]{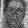} &
\includegraphics[width=1.5cm,height=1.5cm]{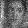} &
\includegraphics[width=1.5cm,height=1.5cm]{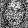} &
\includegraphics[width=1.5cm,height=1.5cm]{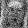} &
\includegraphics[width=1.5cm,height=1.5cm]{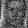} &
\includegraphics[width=1.5cm,height=1.5cm]{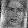} &
\makecell{Not \\ Recognize\\ able} &
\makecell{$0.5{+}0.8{+}0.6{+}0.5{+}0.4{+}0.7{+}0.7{+}0.8{+}0.9{+}0{=}$ \\ \textbf{5.9}} \\
\hline

\makecell[l]{WDIWCD \\ \textbf{(proposed)}} &
\includegraphics[width=1.5cm,height=1.5cm]{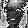} &
\includegraphics[width=1.5cm,height=1.5cm]{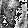} &
\includegraphics[width=1.5cm,height=1.5cm]{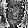} &
\includegraphics[width=1.5cm,height=1.5cm]{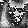} &
\includegraphics[width=1.5cm,height=1.5cm]{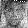} &
\includegraphics[width=1.5cm,height=1.5cm]{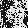} &
\includegraphics[width=1.5cm,height=1.5cm]{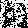} &
\includegraphics[width=1.5cm,height=1.5cm]{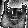} &
\includegraphics[width=1.5cm,height=1.5cm]{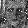} &
\includegraphics[width=1.5cm,height=1.5cm]{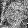} &
\makecell{$0.3{+}0.3{+}0.3{+}0.7{+}0.9{+}0.6{+}0.6{+}0.8{+}0.4{+}0.9{=}$ \\ \textbf{5.8}} \\
\hline

\makecell[l]{VAE-CECL \\ Last Laver} &
\includegraphics[width=1.5cm,height=1.5cm]{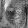} &
\includegraphics[width=1.5cm,height=1.5cm]{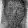} &
\includegraphics[width=1.5cm,height=1.5cm]{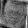} &
\includegraphics[width=1.5cm,height=1.5cm]{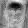} &
\makecell{Not \\ Recognize\\ able} &
\includegraphics[width=1.5cm,height=1.5cm]{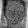} &
\includegraphics[width=1.5cm,height=1.5cm]{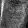} &
\includegraphics[width=1.5cm,height=1.5cm]{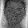} &
\includegraphics[width=1.5cm,height=1.5cm]{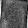} &
\makecell{Not \\ Recognize\\ able} &
\makecell{$0.6{+}0.9{+}0.9{+}0.6{+}0{+}0.9{+}0.6{+}0.8{+}0.3{+}0{=}$ \\ \textbf{5.6}} \\
\hline

NWDDWCC &
\includegraphics[width=1.5cm,height=1.5cm]{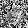} &
\includegraphics[width=1.5cm,height=1.5cm]{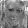} &
\includegraphics[width=1.5cm,height=1.5cm]{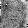} &
\includegraphics[width=1.5cm,height=1.5cm]{images/image46.png} &
\includegraphics[width=1.5cm,height=1.5cm]{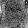} &
\includegraphics[width=1.5cm,height=1.5cm]{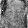} &
\includegraphics[width=1.5cm,height=1.5cm]{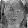} &
\includegraphics[width=1.5cm,height=1.5cm]{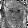} &
\includegraphics[width=1.5cm,height=1.5cm]{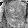} &
\includegraphics[width=1.5cm,height=1.5cm]{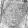} &
\makecell{$0.5{+}0.1{+}0.1{+}0.7{+}0.7{+}0.9{+}0.9{+}0.6{+}0.3{+}0.8{=}$ \\ \textbf{5.6}} \\
\hline

WDDWCC &
\includegraphics[width=1.5cm,height=1.5cm]{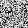} &
\includegraphics[width=1.5cm,height=1.5cm]{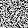} &
\includegraphics[width=1.5cm,height=1.5cm]{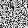} &
\includegraphics[width=1.5cm,height=1.5cm]{images/image101.png} &
\includegraphics[width=1.5cm,height=1.5cm]{images/image101.png} &
\includegraphics[width=1.5cm,height=1.5cm]{images/image101.png} &
\includegraphics[width=1.5cm,height=1.5cm]{images/image101.png} &
\includegraphics[width=1.5cm,height=1.5cm]{images/image101.png} &
\includegraphics[width=1.5cm,height=1.5cm]{images/image101.png} &
\includegraphics[width=1.5cm,height=1.5cm]{images/image101.png} &
\makecell{$0.3{+}0.3{+}0.6{+}0.3{+}0.3{+}0.6{+}0.6{+}0.6{+}0.6{+}0.6{=}$ \\ \textbf{4.8}} \\
\hline

\makecell[l]{PCA (Unsup.) \\ \textbf{(baseline)}} &
\includegraphics[width=1.5cm,height=1.5cm]{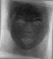} &
\includegraphics[width=1.5cm,height=1.5cm]{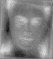} &
\includegraphics[width=1.5cm,height=1.5cm]{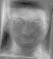} &
\includegraphics[width=1.5cm,height=1.5cm]{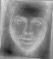} &
\includegraphics[width=1.5cm,height=1.5cm]{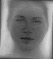} &
\includegraphics[width=1.5cm,height=1.5cm]{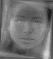} &
\includegraphics[width=1.5cm,height=1.5cm]{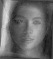} &
\includegraphics[width=1.5cm,height=1.5cm]{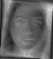} &
\includegraphics[width=1.5cm,height=1.5cm]{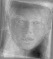} &
\includegraphics[width=1.5cm,height=1.5cm]{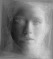} &
\makecell{$0.6{+}0.4{+}0.6{+}0.6{+}0.2{+}0.2{+}0.3{+}0.3{+}0.6{+}0.5{=}$ \\ \textbf{4.3}} \\
\hline

\makecell[l]{VAE-MWCCE \\ First Layer} &
\includegraphics[width=1.5cm,height=1.5cm]{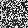} &
\includegraphics[width=1.5cm,height=1.5cm]{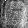} &
\makecell{Not \\ Recognize\\ able} &
\includegraphics[width=1.5cm,height=1.5cm]{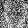} &
\includegraphics[width=1.5cm,height=1.5cm]{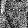} &
\includegraphics[width=1.5cm,height=1.5cm]{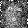} &
\includegraphics[width=1.5cm,height=1.5cm]{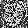} &
\makecell{Not \\ Recognize\\ able} &
\makecell{Not \\ Recognize\\ able} &
\makecell{Not \\ Recognize\\ able} &
\makecell{$0.4{+}0.3{+}0{+}0.3{+}0.4{+}0.6{+}0.5{+}0{+}0{+}0{=}$ \\ \textbf{2.5}} \\
\hline

\makecell[l]{VAE \\ 1st Layer \textbf{(baseline)}} &
\includegraphics[width=1.5cm,height=1.5cm]{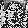} &
\includegraphics[width=1.5cm,height=1.5cm]{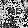} &
\makecell{Not \\ Recognize\\ able} &
\makecell{Not \\ Recognize\\ able} &
\makecell{Not \\ Recognize\\ able} &
\includegraphics[width=1.5cm,height=1.5cm]{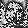} &
\includegraphics[width=1.5cm,height=1.5cm]{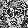} &
\makecell{Not \\ Recognize\\ able} &
\makecell{Not \\ Recognize\\ able} &
\makecell{Not \\ Recognize\\ able} &
\makecell{$0.3{+}0.7{+}0{+}0{+}0{+}0.6{+}0.6{+}0{+}0{+}0{=}$ \\ \textbf{2.2}} \\
\hline

\end{tabular}}
\end{table*}

\begin{table*}[!ht]
\caption{Comparison of all methods layer-shared with a Variational Autoencoder for the MNIST dataset. MNRS = Manual Number Recognition Score (interpretability). WDDWCC = Whole-Data Decorrelation Within-Class Correlation; WDIWCD = Whole-Data Independence Within-Class Dependence; NWDIWCD = Neural WDIWCD; CECL = Class Entities Largest Correlation; MWCCE = Max Within-Class Cross-Entropy; WDMEWCME = Whole-Data Max Entropy Within-Class Min Entropy. Numerical scores are means over five independent runs; `$\wedge$' denotes statistical significance ($p<0.05$, paired $t$-test).}
\centering
\label{tab:table7}
\scriptsize
\setlength{\tabcolsep}{2pt}
\resizebox{\textwidth}{!}{%
\begin{tabular}{|l|*{10}{c|}r|}
\hline
\textbf{Weight parameter} & \textbf{0} & \textbf{1} & \textbf{2} & \textbf{3} & \textbf{4} & \textbf{5} & \textbf{6} & \textbf{7} & \textbf{8} & \textbf{9} & \textbf{MNRS} \\
\hline

Dataset Sample &
\includegraphics[width=1.5cm,height=1.5cm]{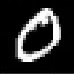} &
\includegraphics[width=1.5cm,height=1.5cm]{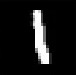} &
\includegraphics[width=1.5cm,height=1.5cm]{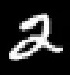} &
\includegraphics[width=1.5cm,height=1.5cm]{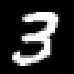} &
\includegraphics[width=1.5cm,height=1.5cm]{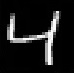} &
\includegraphics[width=1.5cm,height=1.5cm]{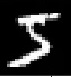} &
\includegraphics[width=1.5cm,height=1.5cm]{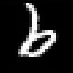} &
\includegraphics[width=1.5cm,height=1.5cm]{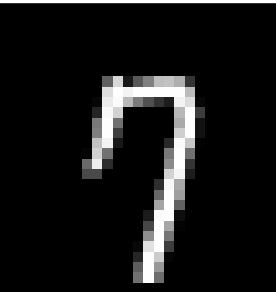} &
\includegraphics[width=1.5cm,height=1.5cm]{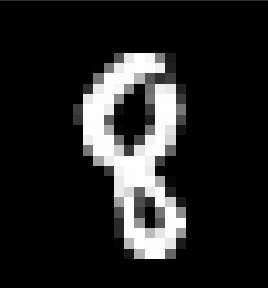} &
\includegraphics[width=1.5cm,height=1.5cm]{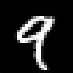} &
\makecell{$1{+}1{+}1{+}1{+}1{+}1{+}1{+}1{+}1{+}1{=}$ \\ \textbf{10}} \\
\hline

\makecell[l]{Neural WDIWCD \\ \textbf{(proposed)}} &
\includegraphics[width=1.5cm,height=1.5cm]{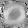} &
\includegraphics[width=1.5cm,height=1.5cm]{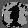} &
\makecell{Not \\ Recognize\\ able} &
\includegraphics[width=1.5cm,height=1.5cm]{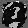} &
\includegraphics[width=1.5cm,height=1.5cm]{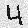} &
\includegraphics[width=1.5cm,height=1.5cm]{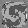} &
\includegraphics[width=1.5cm,height=1.5cm]{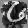} &
\includegraphics[width=1.5cm,height=1.5cm]{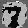} &
\makecell{Not \\ Recognize\\ able} &
\includegraphics[width=1.5cm,height=1.5cm]{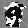} &
\makecell{$0.9{+}0.9{+}0{+}0.5{+}1{+}0.8{+}0.9{+}0.8{+}0.9{+}0{+}0.9{=}$ \\ \textbf{7.6}} \\
\hline

\makecell[l]{VAE \\ Last Layer \\ \textbf{(baseline)}} &
\includegraphics[width=1.5cm,height=1.5cm]{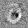} &
\makecell{Not \\ Recognize\\ able} &
\includegraphics[width=1.5cm,height=1.5cm]{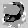} &
\includegraphics[width=1.5cm,height=1.5cm]{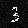} &
\includegraphics[width=1.5cm,height=1.5cm]{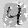} &
\includegraphics[width=1.5cm,height=1.5cm]{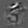} &
\includegraphics[width=1.5cm,height=1.5cm]{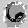} &
\includegraphics[width=1.5cm,height=1.5cm]{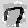} &
\includegraphics[width=1.5cm,height=1.5cm]{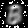} &
\includegraphics[width=1.5cm,height=1.5cm]{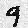} &
\makecell{$0.7{+}0{+}0.6{+}0.9{+}0.8{+}0.6{+}0.8{+}0.8{+}0.6{+}0.9{=}$ \\ \textbf{6.7}} \\
\hline

\makecell[l]{VAE-WDDWCC \\ First Laver \\ \textbf{(proposed)}} &
\includegraphics[width=1.5cm,height=1.5cm]{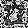} &
\includegraphics[width=1.5cm,height=1.5cm]{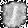} &
\includegraphics[width=1.5cm,height=1.5cm]{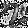} &
\includegraphics[width=1.5cm,height=1.5cm]{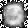} &
\includegraphics[width=1.5cm,height=1.5cm]{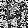} &
\includegraphics[width=1.5cm,height=1.5cm]{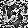} &
\includegraphics[width=1.5cm,height=1.5cm]{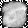} &
\includegraphics[width=1.5cm,height=1.5cm]{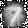} &
\includegraphics[width=1.5cm,height=1.5cm]{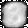} &
\includegraphics[width=1.5cm,height=1.5cm]{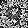} &
\makecell{$0.6{+}0.7{+}0.8{+}0.8{+}0.6{+}0.6{+}0.6{+}0.7{+}0.7{+}0.5{=}$ \\ \textbf{6.6}} \\
\hline

\makecell[l]{VAE-CELC \\ 1st Layer} &
\includegraphics[width=1.5cm,height=1.5cm]{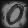} &
\includegraphics[width=1.5cm,height=1.5cm]{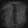} &
\includegraphics[width=1.5cm,height=1.5cm]{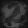} &
\includegraphics[width=1.5cm,height=1.5cm]{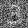} &
\includegraphics[width=1.5cm,height=1.5cm]{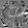} &
\includegraphics[width=1.5cm,height=1.5cm]{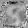} &
\includegraphics[width=1.5cm,height=1.5cm]{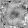} &
\includegraphics[width=1.5cm,height=1.5cm]{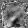} &
\includegraphics[width=1.5cm,height=1.5cm]{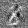} &
\includegraphics[width=1.5cm,height=1.5cm]{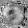} &
\makecell{$0.9{+}0.8{+}0.7{+}0.6{+}0.6{+}0.5{+}0.6{+}0.6{+}0.7{+}0.6{=}$ \\ \textbf{6.6}} \\
\hline

\makecell[l]{VAE \\ 1st Layer \\ \textbf{(baseline)}} &
\includegraphics[width=1.5cm,height=1.5cm]{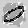} &
\makecell{Not \\ Recognize\\ able} &
\includegraphics[width=1.5cm,height=1.5cm]{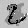} &
\includegraphics[width=1.5cm,height=1.5cm]{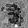} &
\includegraphics[width=1.5cm,height=1.5cm]{images_b/image22.png} &
\includegraphics[width=1.5cm,height=1.5cm]{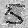} &
\includegraphics[width=1.5cm,height=1.5cm]{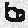} &
\includegraphics[width=1.5cm,height=1.5cm]{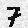} &
\includegraphics[width=1.5cm,height=1.5cm]{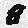} &
\includegraphics[width=1.5cm,height=1.5cm]{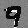} &
\makecell{$0.7{+}0{+}0.6{+}0.5{+}0.8{+}0.8{+}0.5{+}0.8{+}0.8{+}0.8{=}$ \\ \textbf{6.3}} \\
\hline

\makecell[l]{NWDDWCC \\ \textbf{(proposed)}} &
\includegraphics[width=1.5cm,height=1.5cm]{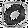} &
\includegraphics[width=1.5cm,height=1.5cm]{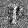} &
\makecell{Not \\ Recognize\\ able} &
\includegraphics[width=1.5cm,height=1.5cm]{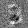} &
\includegraphics[width=1.5cm,height=1.5cm]{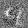} &
\includegraphics[width=1.5cm,height=1.5cm]{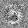} &
\includegraphics[width=1.5cm,height=1.5cm]{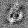} &
\includegraphics[width=1.5cm,height=1.5cm]{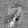} &
\includegraphics[width=1.5cm,height=1.5cm]{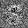} &
\includegraphics[width=1.5cm,height=1.5cm]{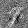} &
\makecell{$1{+}0.7{+}0{+}0.6{+}0.7{+}0.5{+}0.6{+}0.7{+}0.6{+}0.5{=}$ \\ \textbf{5.9}} \\
\hline

\makecell[l]{VAE-WDDWCC \\ Last Layer \\ \textbf{(proposed)}} &
\includegraphics[width=1.5cm,height=1.5cm]{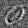} &
\makecell{Not \\ Recognize\\ able} &
\includegraphics[width=1.5cm,height=1.5cm]{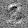} &
\includegraphics[width=1.5cm,height=1.5cm]{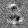} &
\includegraphics[width=1.5cm,height=1.5cm]{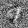} &
\includegraphics[width=1.5cm,height=1.5cm]{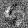} &
\includegraphics[width=1.5cm,height=1.5cm]{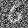} &
\includegraphics[width=1.5cm,height=1.5cm]{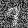} &
\includegraphics[width=1.5cm,height=1.5cm]{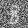} &
\includegraphics[width=1.5cm,height=1.5cm]{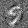} &
\makecell{$0.7{+}0{+}0.6{+}0.5{+}0.5{+}0.6{+}0.7{+}0.5{+}0.7{+}0.5{=}$ \\ \textbf{5.3}} \\
\hline

\makecell[l]{VAE-WDIWCD \\ First layer \\ \textbf{(proposed)}} &
\includegraphics[width=1.5cm,height=1.5cm]{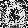} &
\includegraphics[width=1.5cm,height=1.5cm]{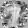} &
\includegraphics[width=1.5cm,height=1.5cm]{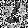} &
\includegraphics[width=1.5cm,height=1.5cm]{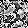} &
\includegraphics[width=1.5cm,height=1.5cm]{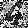} &
\includegraphics[width=1.5cm,height=1.5cm]{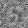} &
\includegraphics[width=1.5cm,height=1.5cm]{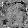} &
\includegraphics[width=1.5cm,height=1.5cm]{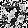} &
\includegraphics[width=1.5cm,height=1.5cm]{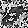} &
\includegraphics[width=1.5cm,height=1.5cm]{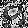} &
\makecell{$0.3{+}0.6{+}0.6{+}0.7{+}0.8{+}0.2{+}0.1{+}0.8{+}0.3{+}0.4{=}$ \\ \textbf{4.8}} \\
\hline

\makecell[l]{VAE-WDIWCD \\ Last layer \\ \textbf{(proposed)}} &
\includegraphics[width=1.5cm,height=1.5cm]{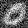} &
\makecell{Not \\ Recognize\\ able} &
\makecell{Not \\ Recognize\\ able} &
\includegraphics[width=1.5cm,height=1.5cm]{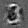} &
\includegraphics[width=1.5cm,height=1.5cm]{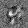} &
\includegraphics[width=1.5cm,height=1.5cm]{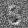} &
\includegraphics[width=1.5cm,height=1.5cm]{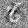} &
\makecell{Not \\ Recognize\\ able} &
\includegraphics[width=1.5cm,height=1.5cm]{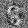} &
\includegraphics[width=1.5cm,height=1.5cm]{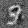} &
\makecell{$0.8{+}0{+}0{+}0.9{+}0.1{+}0.6{+}0.7{+}0{+}0.5{+}0.9{=}$ \\ \textbf{4.5}} \\
\hline

\makecell[l]{VAE-CELC \\ Last Laver} &
\includegraphics[width=1.5cm,height=1.5cm]{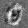} &
\makecell{Not \\ Recognize\\ able} &
\includegraphics[width=1.5cm,height=1.5cm]{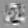} &
\makecell{Not \\ Recognize\\ able} &
\makecell{Not \\ Recognize\\ able} &
\includegraphics[width=1.5cm,height=1.5cm]{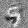} &
\includegraphics[width=1.5cm,height=1.5cm]{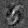} &
\includegraphics[width=1.5cm,height=1.5cm]{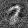} &
\includegraphics[width=1.5cm,height=1.5cm]{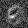} &
\includegraphics[width=1.5cm,height=1.5cm]{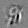} &
\makecell{$0.5{+}0{+}0.7{+}0{+}0{+}0.6{+}0.6{+}0.7{+}0.5{+}0.7{=}$ \\ \textbf{4.3}} \\
\hline

\makecell[l]{VAE-WDMEWCME \\ 1st Laver \\ \textbf{(proposed)}} &
\includegraphics[width=1.5cm,height=1.5cm]{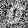} &
\makecell{Not \\ Recognize\\ able} &
\makecell{Not \\ Recognize\\ able} &
\makecell{Not \\ Recognize\\ able} &
\makecell{Not \\ Recognize\\ able} &
\includegraphics[width=1.5cm,height=1.5cm]{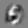} &
\includegraphics[width=1.5cm,height=1.5cm]{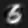} &
\includegraphics[width=1.5cm,height=1.5cm]{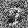} &
\includegraphics[width=1.5cm,height=1.5cm]{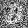} &
\includegraphics[width=1.5cm,height=1.5cm]{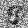} &
\makecell{$0.6{+}0{+}0{+}0{+}0{+}0.8{+}0.6{+}0.6{+}0.5{+}0.4{=}$ \\ \textbf{3.5}} \\
\hline

\makecell[l]{RLDA \\ \textbf{(baseline)}} &
\includegraphics[width=1.5cm,height=1.5cm]{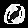} &
\makecell{Not \\ Recognize\\ able} &
\makecell{Not \\ Recognize\\ able} &
\makecell{Not \\ Recognize\\ able} &
\includegraphics[width=1.5cm,height=1.5cm]{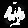} &
\makecell{Not \\ Recognize\\ able} &
\makecell{Not \\ Recognize\\ able} &
\includegraphics[width=1.5cm,height=1.5cm]{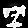} &
\makecell{Not \\ Recognize\\ able} &
\includegraphics[width=1.5cm,height=1.5cm]{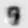} &
\makecell{$1{+}0{+}0{+}0{+}0.7{+}0{+}0{+}0.8{+}0{+}1{=}$ \\ \textbf{3.5}} \\
\hline

\makecell[l]{PCA (Unsup.) \\ \textbf{(baseline)}} &
\includegraphics[width=1.5cm,height=1.5cm]{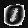} &
\includegraphics[width=1.5cm,height=1.5cm]{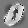} &
\makecell{Not \\ Recognize\\ able} &
\includegraphics[width=1.5cm,height=1.5cm]{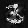} &
\includegraphics[width=1.5cm,height=1.5cm]{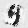} &
\includegraphics[width=1.5cm,height=1.5cm]{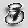} &
\makecell{Not \\ Recognize\\ able} &
\makecell{Not \\ Recognize\\ able} &
\makecell{Not \\ Recognize\\ able} &
\makecell{Not \\ Recognize\\ able} &
\makecell{$0.4{+}0.4{+}0{+}0.9{+}0.9{+}0.6{+}0{+}0{+}0{+}0{=}$ \\ \textbf{3.2}} \\
\hline

\makecell[l]{Null-space \\ Decorrelation} &
\includegraphics[width=1.5cm,height=1.5cm]{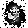} &
\makecell{Not \\ Recognize\\ able} &
\includegraphics[width=1.5cm,height=1.5cm]{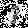} &
\includegraphics[width=1.5cm,height=1.5cm]{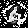} &
\includegraphics[width=1.5cm,height=1.5cm]{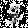} &
\includegraphics[width=1.5cm,height=1.5cm]{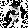} &
\makecell{Not \\ Recognize\\ able} &
\makecell{Not \\ Recognize\\ able} &
\makecell{Not \\ Recognize\\ able} &
\includegraphics[width=1.5cm,height=1.5cm]{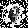} &
\makecell{$0.8{+}0{+}0.4{+}0.3{+}0.4{+}0.5{+}0{+}0{+}0{+}0.3{=}$ \\ \textbf{2.7}} \\
\hline

\makecell[l]{VAE-MWCCE \\ First Layer} &
\makecell{Not \\ Recognize\\ able} &
\makecell{Not \\ Recognize\\ able} &
\includegraphics[width=1.5cm,height=1.5cm]{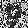} &
\makecell{Not \\ Recognize\\ able} &
\makecell{Not \\ Recognize\\ able} &
\makecell{Not \\ Recognize\\ able} &
\makecell{Not \\ Recognize\\ able} &
\makecell{Not \\ Recognize\\ able} &
\includegraphics[width=1.5cm,height=1.5cm]{images_b/image82.png} &
\includegraphics[width=1.5cm,height=1.5cm]{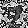} &
\makecell{$0{+}0{+}0.1{+}0{+}0{+}0{+}0{+}0{+}0.6{+}0.6{=}$ \\ \textbf{1.3}} \\
\hline

\end{tabular}}
\end{table*}

\begin{figure}[htbp]
  \centering
  \includegraphics[width=\columnwidth]{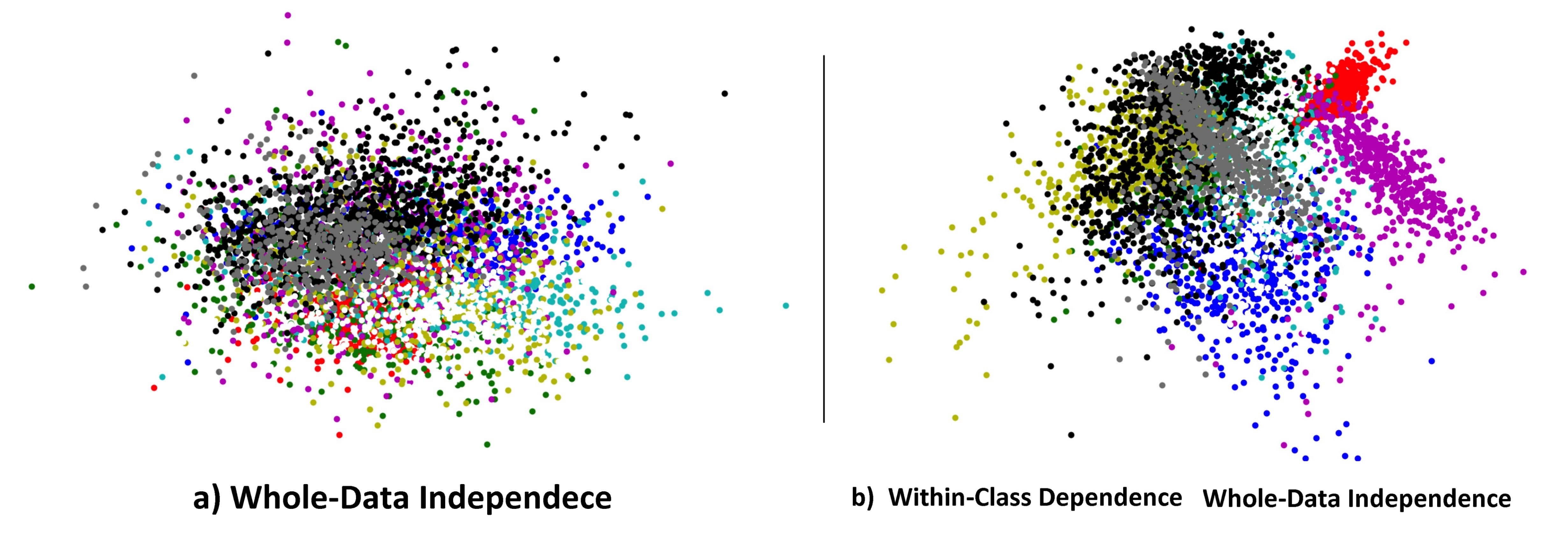}
  \caption{Comparison of LIBCA and WDIWCD. LIBCA produces wider, more shuffled projections, while WDIWCD produces more class-distinct clusters.}
  \label{fig:libcawdiwcd}
\end{figure}

From the results in Tables~\ref{tab:table6} and~\ref{tab:table7}, the following observations emerge:
\begin{itemize}
  \item The proposed supervised feature extraction methods produce more class-discriminative eigenfaces and eigenimages, and further improve when combined with nonlinear models such as VAE.
  \item The algorithm occasionally generates patterns resembling two digits simultaneously; once the ambiguity is resolved during training, the cost ceases to decrease optimally. This behavior, absent in other layers, suggests an energy-efficiency mechanism in the shared layer.
  \item Shared layer weights appear to trigger implicit data augmentation, with some digits appearing rotated or inverted.
  \item The whole-data independence criterion captures between-sample structure while learning the optimal histogram, acting like a kernel that conveys holistic topological information to the first VAE layer---similar in effect to pretraining.
  \item Because NWDIWCD backpropagates through more layers than its linear counterpart, it controls a richer hierarchy of features, including fine patterns for recognizing submodules such as noses or lips, which explains why circles appear instead of lip shapes in some eigenfaces.
  \item VAE alone produces fewer distinct eigenimages; its performance and diversity are maximized when combined with WDIWCD or WDDWCC. Fine-tuning reveals that during hybridization the weight $c$ is much larger than $1-c$ in~(\ref{eq:28}), meaning the whole-data independence term contributes more heavily to the shared layer.
  \item For class discrimination, the best configuration combines WDIWCD with VAE.
  \item Table~\ref{tab:table8} confirms that layer sharing yields higher accuracy and lower Mean Squared Error (MSE) than VAE alone.
\end{itemize}

Results are averaged over 10 runs. The optimal hyperparameters from~(\ref{eq:28}) were $a=0.8$, $b=0.48$, $c=0.87$.

Statistical comparisons in Table~\ref{tab:table8} are based on a paired $t$-test over five independent runs (significance level $\alpha=0.05$). Each run uses a different random seed for weight initialization; the paired structure matches each proposed method run against the corresponding baseline run under the same seed, controlling for initialization variance. The symbol `$\wedge$' in the table marks methods that achieve statistically significant improvements over the standalone VAE baseline. 

\textbf{RQ3 validation.}
In Table~\ref{tab:table8}, VAE+WDIWCD achieves 89.4\% on MNIST-2 and 92.7\% on Gender, outperforming standalone WDIWCD (88.9\%, 90.9\%) and standalone VAE (84.2\%, 81.6\%). VAE+WDDWCC similarly achieves 89.6\%/78.6\%/90.1\% across three accuracy columns, consistently exceeding standalone WDDWCC (87.5\%/71.0\%/89.1\%). Layer sharing outperforms the standalone version in 3/4 columns for VAE+WDIWCD and 4/4 columns for VAE+WDDWCC, confirming a systematic benefit (RQ3). WDIWCD achieves higher accuracy than WDDWCC in 2 of 4 columns, supporting dependence as a beneficial inductive bias for classification (RQ2).

\begin{table}[htbp]
\caption{KNN classification accuracy (\%) and VAE mean squared error (MSE) for the proposed feature extractions. All values are mean~$\pm$~standard deviation over five independent runs. Statistical significance is assessed by a paired $t$-test against the standalone VAE baseline ($\alpha=0.05$); `$\wedge$' marks statistically significant results. SPC is Supervised Principal Components. The proposed DR methods outperform supervised and unsupervised DR.}
\label{tab:table8}
\resizebox{\columnwidth}{!}{%
\begin{tabular}{|l|c|c|c|c|}
\hline
 & \textbf{Acc2 MNIST} & \textbf{Acc10 MNIST} & \textbf{Acc2 Gender} & \textbf{MSE} \\
\hline
VAE+WDIWCD & $\mathbf{89.4{\pm}2.09^{\wedge}}$ & $\mathbf{75.5{\pm}2.70^{\wedge}}$ & $\mathbf{92.7{\pm}4.47^{\wedge}}$ & $\mathbf{0.019{\pm}0.0016^{\wedge}}$ \\
\hline
Only WDIWCD & $88.9{\pm}1.89^{\wedge}$ & $72.6{\pm}1.53^{\wedge}$ & $\mathbf{90.9{\pm}3.75^{\wedge}}$ & $\mathbf{0.016{\pm}0.0040^{\wedge}}$ \\
\hline
Only VAE & 84.2 & 68.8 & 81.6 & 0.021 \\
\hline
VAE+WDDWCC & $\mathbf{89.6{\pm}2.18^{\wedge}}$ & $\mathbf{78.6{\pm}3.95^{\wedge}}$ & $90.1{\pm}3.42^{\wedge}$ & $0.020{\pm}0.0008^{\wedge}$ \\
\hline
Only WDDWCC & $87.5{\pm}1.33^{\wedge}$ & $71.0{\pm}0.89^{\wedge}$ & $89.1{\pm}3.02^{\wedge}$ & $0.023{\pm}0.0016^{\wedge}$ \\
\hline
VAE+CEL & $85.0{\pm}0.32^{\wedge}$ & $75.1{\pm}4.60$ & $83.9{\pm}0.93^{\wedge}$ & $0.023{\pm}0.0016^{\wedge}$ \\ 
\hline
Only CEL & $83.4{\pm}0.32^{\wedge}$ & $73.8{\pm}2.02^{\wedge}$ & $82.8{\pm}0.88$ & $0.025{\pm}0.0032^{\wedge}$ \\
\hline
LDA & $81.2{\pm}0.20^{\wedge}$ & $69.7{\pm}1.6^{\wedge}$ & $81.2{\pm}0.3$ & $0.010{\pm}0.0017^{\wedge}$ \\
\hline
SPC & $81.5{\pm}0.1^{\wedge}$ & $71.4{\pm}1.6^{\wedge}$ & $81.8{\pm}0.5$ & $0.019{\pm}0.0015^{\wedge}$ \\
\hline
\end{tabular}}
\end{table}

\section{Conclusion and Future Work}
\label{sec:conclusion}

In this work, we investigated dependence and independence criteria for supervised and unsupervised DR and proposed new dependence-driven learning objectives for discriminative and interpretable representation learning. The proposed framework combines null-space-correlation and null-space-independence criteria with class-objectives and further extends these ideas through neural architectures and layer sharing with VAEs. In addition to classification and reconstruction performance, we analyzed eigenimages and eigenfaces to study the interpretability and data-contrast structure of the learned representations.

Experimental results on MNIST and gender-face datasets demonstrated consistent improvements over PCA, LDA, ICA variants, t-SNE, and VAE baselines. In particular, the proposed methods improved contrast quality by up to 20.1\% over PCA, classification accuracy by 17.4\% over LDA and 14.2\% over VAE, and visual interpretability scores by up to 120\% over Regularized LDA. Layer sharing with VAE additionally reduced reconstruction error by 9.5\%, demonstrating that dependence-aware objectives can simultaneously improve latent representation quality and reconstruction.

The three research questions are addressed as follows. RQ1 is validated by improved ICA source-separation results (Table~\ref{tab:table3}, Amari index 1.5 vs.\ 4.4--5.8), positive contrast results (Table~\ref{tab:table2}, Figure~\ref{fig:contrast}), and positive class-separability (Figure~\ref{fig:classsep}). RQ2 is confirmed by consistent 20--38\% interpretability gains of WDIWCD over WDDWCC across Tables~\ref{tab:table6}--\ref{tab:table7}, demonstrating that replacing correlation with dependence improves class-specific interpretability, and by KNN accuracy improvements in Table~\ref{tab:table8}. RQ3 is supported by the systematic outperformance of VAE-combined methods over their standalone counterparts in both accuracy (Table~\ref{tab:table8}) and interpretability (Tables~\ref{tab:table6}--\ref{tab:table7}).

Future work will focus on hierarchical softmax target encoding and more scalable dependence estimation methods. Extensions to disentangled and self-supervised representation learning settings are also planned, as recent progress in contrastive and self-supervised learning~\cite{ref26,ref27,ref29} offers a natural complement to the proposed dependence-based objectives. Overall, the results indicate that dependence and independence criteria provide a principled and practically effective direction for interpretable and discriminative DR.

\end{document}